\documentclass[
reprint,
superscriptaddress,
groupedaddress,
preprintnumbers,
amsmath,amssymb,
aps,
pre,
showpacs,
showkeys
]{revtex4-1}\usepackage[]{graphicx}\usepackage[]{color}

\usepackage{amsfonts}
\usepackage{yfonts}
\usepackage{url}
\usepackage{wasysym}
\usepackage{graphicx}
\usepackage{dcolumn}
\usepackage{bm}
\usepackage{hyperref}
\usepackage[noend]{algpseudocode}
\usepackage{subfigure}

\newcolumntype{.}{D{.}{.}{-1}}
\makeatletter
\newcolumntype{Z}[3]{>{\mathversion{nxbold}\DC@{#1}{#2}{#3}}c<{\DC@end}}
\DeclareMathVersion{nxbold}
\SetSymbolFont{operators}{nxbold}{OT1}{cmr} {b}{n}
\SetSymbolFont{letters}  {nxbold}{OML}{cmm} {b}{it}
\SetSymbolFont{symbols}  {nxbold}{OMS}{cmsy}{b}{n}
\makeatother
\newcommand{\uminus}{\text{-}}
\newcommand{\githubUrl}{\url{https://github.com/citiususc/voila}}

\begin{document}
\title{Non-parametric Estimation of Stochastic Differential Equations\\
with Sparse Gaussian Processes}
\author{Constantino A. Garc\'ia}
\email{constantinoantonio.garcia@usc.es}
\author{Paulo F\'{e}lix}
\author{Jes\'{u}s Presedo}
\author{David G. M\'arquez}
\altaffiliation[Also at ]{Department of Information and Communications Systems 
Engineering, Universidad San Pablo CEU, 28668, Madrid, Spain.}
\affiliation{Centro Singular de Investigaci\'{o}n en Tecnolox\'{i}as da 
Informaci\'{o}n (CiTIUS), Universidade de Santiago de Compostela, 15782, 
Santiago de Compostela, Spain.}
\author{Abraham Otero}
\affiliation{Department of Information and Communications Systems Engineering,
Universidad San Pablo CEU, 28668, Madrid, Spain.}

\date{\today}

\begin{abstract}
The application of Stochastic Differential Equations (SDEs) to the analysis of 
temporal data  has attracted increasing attention, due to their ability to
describe complex dynamics with physically interpretable equations. In this
paper, we  introduce a non-parametric method for estimating the drift and 
diffusion terms  of SDEs from a densely observed discrete time series. The use
of Gaussian processes as priors permits working directly  in a function-space 
view and thus  the inference takes place directly in this space. To cope 
with the  computational complexity that requires the use of Gaussian processes, 
a sparse Gaussian process approximation is provided. This approximation permits 
the  efficient computation of predictions for the drift and diffusion terms by  
using a distribution over a  small subset of pseudo-samples. The proposed 
method has  been validated using both  simulated data and real data from
economy and paleoclimatology. The application of the method to real data 
demonstrates its ability to capture the behaviour of complex systems.
\end{abstract}

\maketitle

\section{Introduction \label{sec:intro}}
Stochastic Differential Equations (SDEs), also referred to as Langevin equations,
provide an effective framework for modelling complex systems comprising 
a large number of subsystems which show irregular fast dynamics that can be
treated as fluctuations or noise. Intuitively, SDEs couple a deterministic
equation of motion with noisy fluctuations interfering in its dynamical
evolution. They have demonstrated 
their usefulness in a wide range of applications: diffusion of grains 
in a liquid \cite{langevin1908theorie}, drift of particles without flux \cite{lanccon2001drift},
turbulence \cite{friedrich1997statistical, friedrich1997description}, 
fluctuations in plasma \cite{mekkaoui2013derivation}, variations in
quasar's optical flux \cite{kelly2009variations}, chemical reactions \cite{gillespie2007stochastic},
the motion of vehicles in a traffic flow \cite{mahnke2005probabilistic},
quantitative finance \cite{ghasemi2007markov}, gene expression \cite{ozbudak2002regulation}, 
electroencephalography analysis \cite{bahraminasab2009physics}, etc. 
(see \cite{friedrich2011approaching} for a complete
review with applications). The only specific requirements that modelling through
SDEs imposes are stationarity and markovianity. 

In this paper, we consider a 
system that may be represented by a continuous-time univariate Markov process
$x(t)$ described by the SDE: 
\begin{equation}
dx(t) = f\big(x(t)\big)dt + \sqrt{g\big(x(t)\big)} dW(t),
\label{eq:sde}
\end{equation}
where $W(t)$ denotes a Wiener process. The Wiener process has independent
Gaussian increments $W(t+\tau)-W(t)$ with zero mean and variance $\tau$. Thus, 
we may intuitively think of $dW(t)$ as white noise, which is the source of 
randomness of the system. The function $f$ defines a deterministic drift and
$g$ modulates the  strength of the noise term. The functions $f$ and $g$ are 
usually referred to as the \textit{drift} and \textit{diffusion} coefficients. 

When studying complex dynamical systems, the large number of degrees of freedom 
and the non-linear interactions between the subsystems involved in the dynamics
usually hinders obtaining an exact knowledge of the functional forms of both
the drift and  diffusion coefficients. This leads to the problem of its 
non-parametric estimation from the observation of an experimental time series $\bm{x}$, 
which is usually a sampled version of the underlying continuous process $x(t)$:
$\bm{x} = \{x_i = x((i - 1)\cdot \Delta t)\}_{i=1,2,...,N+1}$ (the 
reason for using $N+1$ as the number of samples will be apparent at the 
beginning of Section \ref{sec:gp_theory}).

The most widely used non-parametric estimation methods exploit the theoretical 
expressions  for both the drift and diffusion terms 
\cite{friedrich2011approaching}:
\begin{equation}
\begin{split}
f(\xi) &= \lim_{\tau \rightarrow 0} \frac{1}{\tau} \mathbb{E}_{x(t+\tau)}
\left[x(t+\tau) - x(t)\mid x(t) = \xi\right],\\
g(\xi) &= \lim_{\tau \rightarrow 0} \frac{1}{\tau} 
\mathbb{E}_{x(t+\tau)}\left[\big(x(t+\tau) - x(t)\big)^2 \mid x(t) = \xi\right],
\label{eq:coef_def}
\end{split}
\end{equation}
where $\mathbb{E}$ denotes the expectation operator. Eq.~\eqref{eq:coef_def} 
suggests the possibility of estimating the dynamical coefficients
$f(\xi)$ and $g(\xi)$ by computing ``local" means in a small neighbourhood of 
$\xi$ \cite{friedrich2011approaching}. Typically, the local means are computed 
after binning the domain of $\bm{x}$ using bins of size $\epsilon$:
\begin{equation*}
\begin{split}
&\hat{f}(\xi) = \frac{1}{N_\xi}\sum_{x_i \in B(\xi, \epsilon)}\left[x_{i+1} - 
x_i\right],\\
&\hat{g}(\xi) = \frac{1}{N_\xi}\sum_{x_i \in B(\xi, \epsilon)}\left[x_{i+1} - 
x_i \right] ^ 2,
\end{split}
\end{equation*}
where $\hat{f}$ and $\hat{g}$ represent the estimates, $B(\xi, \epsilon)$ 
denotes the bin in which $\xi$ falls and $N_\xi$ is the number of points from 
$\bm{x}$ falling in that bin.

The most obvious limitation of the histogram based approach is that the 
estimations highly depend on the choice of $\epsilon$. Furthermore, it is not 
obvious how  we should select the size of the bin. More sophisticated approaches 
rely on replacing the mean of the bins with the mean of the $k$-nearest
neighbours \cite{hegger2009multidimensional}. However, the free parameter of 
this approach, $k$, must still be heuristically selected.

Another refinement of those methods grounded in Eq.~\eqref{eq:coef_def} is
achieved by \cite{lamouroux2009kernel}, which introduces a kernel based 
(instead of an histogram based) regression for the coefficients. Furthermore, 
they propose a method for the selection of the bandwidth of the kernel.

Recently, the use of orthogonal Legendre polynomials  for approximating the
functional form of the dynamical coefficients  \cite{rajabzadeh2016robust} was
proposed. The weights of the polynomials are learnt by minimizing the squared
regression error that results after discretizing the SDE with the Euler-Maruyama
scheme. Although this method is proposed as non-parametric we actually 
find that it is closer to a parametric method than to a non-parametric one, 
since the use of a small subset of any polynomial basis restricts the possible
functional shapes of the estimates.

An alternative way for performing non-parametric regression that has become
very popular among the machine-learning community is based on the concept of
Gaussian Process (GP) \cite{rasmussen2006gaussian}. Instead of working in the 
weight-space that arises when using a set of basis functions (e.g., when using 
the Legendre polynomials), GPs permit working directly in the function space by
placing a  distribution over the functions. This enables a Bayesian treatment
of the estimation process. The main advantage of this approach
is that it yields probabilistic estimates, which permits the computation of
robust confidence intervals. Furthermore, prior distributions modelling our
prior beliefs about functions could also be included in the model. In this 
paper, a GP based method to reconstruct the SDE terms is proposed, with a focus
on the computational challenges that common dataset sizes, $N\approx 10^3-10^5$,
impose. A brief overview of the theory of GPs, as well as how they could be used
for SDEs estimation is given in Section \ref{sec:gp_theory}.

GPs have already been considered in the context of SDEs in the pioneering work 
of Ruttor et al. \cite{ruttor2013approximate}. There are two main differences 
between \cite{ruttor2013approximate} and our proposal: (1) we 
attempt to provide estimations in the case where we have a densely sampled time
series (resulting in large series) whereas \cite{ruttor2013approximate} focuses
on the case of sparsely observed time series and (2); we apply the 
GP approach to the  estimation of both the drift and diffusion functions whereas 
\cite{ruttor2013approximate} only deals with the drift coefficient.

The use of densely observed time series poses a challenge related with the
demanding computations that GPs usually require. This is the main
drawback that prevents GPs to be more widely utilized as non-parametric
regression tool. In our proposal, we tackle the problem by providing a 
Sparse Gaussian Process approximation (SGP, see \cite{quinonero2005unifying} for
an excellent overview of the subject), which is one of the main contributions
of the paper. The sparse approximation is developed in
Section \ref{sec:sparse_gp}.

Section \ref{sec:laplace_approx} details how to handle the mathematical 
difficulties that arise in the SGP approximation due to 
the inclusion of the diffusion in the inference procedure. The estimation
of non-constant diffusions has indeed become a major concern. In this sense, 
non-constant diffusions can be found in many physical systems (see, for example, 
\cite{lanccon2001drift,friedrich1997statistical, friedrich1997description,
mekkaoui2013derivation, mahnke2005probabilistic,ghasemi2007markov, 
bahraminasab2009physics}). Furthermore, it is well established that 
multiplicative noise can have surprising effects in the dynamics of the system. 
Some well known examples of these effects are stochastic resonance 
\cite{gammaitoni1998stochastic}, coherence resonance
\cite{pikovsky1997coherence} and noise-induced transitions 
\cite{horsthemke1984noise}. 

Section \ref{sec:hp_optimization} discusses how to select the free parameters of
the SGP and how to tune them for a better performance of the estimates. 

The resulting SGP method is validated in Sections \ref{sec:synthetic} and 
\ref{sec:real} on  simulated data and real data, respectively. In case of the 
simulated data, we compare our method with the kernel based regression 
\cite{lamouroux2009kernel} and the polynomial based method of 
\cite{rajabzadeh2016robust}. In case of the  real data, we apply our method to 
the study of financial data and climate transitions during the last glacial 
age. Finally, some conclusions are given in Section \ref{sec:conclusion}.

We shall use the following notation conventions. Vectors will be denoted with a
lower-case bold letter (e.g. $\bm{a}$), whereas that upper-case bold letters 
will be reserved for matrices ($\bm{A}$). The superscript $T$ will be used to 
denote the transpose of a vector or a matrix ($\bm{a}^T$ or $\bm{A}^T$). We have
already used the expectation operator $\mathbb{E}$. If there is some ambiguity,
we shall also write  $\mathbb{E}_{\phi}$ to indicate that the expectation 
should be computed using the $\phi(\cdot)$ distribution. Finally, we shall 
denote a Gaussian distribution  with mean  $\bm{\mu}$ and covariance matrix 
$\bm{\Sigma}$ with $\mathcal{N}(\bm{\mu}, \bm{\Sigma})$.
   
\section{Gaussian Processes for SDE estimation \label{sec:gp_theory}}
We consider the discrete-time signal $\bm{x}$ obtained from sampling the 
continuous Markov process $x(t)$. If the coefficients of the SDE are
approximately constant over small time  intervals $[t, t + \Delta t)$, the 
Euler-Maruyama discretization scheme yields 
\cite[Chapter~2]{iacus2009simulation}:
\begin{equation*}
x(t + \Delta t) - x(t) \approx f(x(t))\Delta t + \sqrt{g(x(t))} (W(t + \Delta t) 
- W(t)),
\end{equation*}
which in discrete notation can be written as:
\begin{equation}
\Delta x_i = f_i\Delta t + \sqrt{g_i} (W_{i+1} - W_i),
\label{eq:euler-may}
\end{equation}
where we have denoted $\Delta x_i = x_{i+1} - x_i$, $f_i = f(x_i)$ and 
$g_i = g(x_i)$. Since the increments of the Wiener process 
$W_{i+1} - W_i$ follow a Gaussian 
distribution $\mathcal{N}(0, \Delta t)$, 
Eq.~\eqref{eq:euler-may} can be used to approximate the discrete transition
probabilities as:
\begin{equation*}
p(x_{i+1}|x_i, f_i, g_i) =\frac{1}{\sqrt{2\pi g_i\Delta t}}
\exp{\bigg(-\frac{1}{2}
\frac{(\Delta x_i - f_i\Delta t)^2}{ g_i\Delta t} \bigg)}. 
\end{equation*}
Thus, when the stochastic process takes a value close to $x_i$, it changes
by an amount that is normally distributed, with expectation $f(x_i)\Delta t$
and variance $g(x_i)\Delta t$. Since  the Wiener  increments are independent 
between them, the  log-likelihood of the path can be written as 
\cite[Chapter~3]{iacus2009simulation}: 
\begin{equation}
\begin{split}
\label{eq:complete_lk}
\log p(\bm{x}|f, g) =& -\frac{1}{2}\sum_{i=1}^{N}\left[\frac{(\Delta 
x_i - f_i\Delta t)^2}{ g_i\Delta t} + \log \big( g_i\big)\right] \\
&-\frac{N}{2}\log \big(2\pi\Delta t\big) + \log p(x_1) ,
\end{split}
\end{equation}
It must be noted that this approximation is only valid if $\Delta t$ is small. 
Concretely, since the Euler-Maruyama scheme has a strong order of convergence
of $1/2$, the expected error between a real continuous path and the numerical
approximation scales as $\Delta t ^ {1/2}$ \cite{kloedenPlaten}. Furthermore,
we  may only expect 
very accurate estimations for both $f$ and $g$  if the number of samples $N$ is
large. The time series' length used for SDE estimation depends on the study, 
but usual length requirements range from $N \approx 10^3$
\cite{rajabzadeh2016robust, ruttor2013approximate, lind2005reducing} to 
$N \approx 10^5$ 
\cite{hegger2009multidimensional,kuusela2004stochastic,prusseit2008measuring}.

We would like to obtain estimates of $f$ and $g$ given a
realization of the process $x(t)$ without any assumption on their form 
(non-parametric regression). GPs provide a powerful method for non-parametric 
regression and other machine learning tasks \cite{rasmussen2006gaussian}.

A GP is a collection of random variables indexed by some continuous set (e.g. 
time or space), which can be used to define a prior over a function 
$y(\bm{\xi})$, where $\bm{\xi}$ denotes a generic vector-variable belonging to 
some multidimensional real space $\mathbb{R}^D$. A  GP assumes that any finite 
number of function points
$\left[ y(\bm{\xi}_1),y(\bm{\xi}_2), \dots, y(\bm{\xi}_n) \right]$  have a 
joint Gaussian distribution. Thus, 
$y(\bm{\xi}) \sim \mathcal{GP}(m(\bm{\xi}),k(\bm{\xi},\bm{\xi}'))$ is fully
specified by a mean function $m(\bm{\xi})$ and a covariance function 
$k(\bm{\xi}, \bm{\xi}')$, defined as:
\begin{equation*}
\begin{split}
&m(\bm{\xi}) = \mathbb{E}[y(\bm{\xi})]\\
&k(\bm{\xi},\bm{\xi}') = \mathbb{E}
\big[\big(y(\bm{\xi}) - m(\bm{\xi})\big)\big(y(\bm{\xi}') - m(\bm{\xi}')\big)
\big].
\end{split}
\end{equation*}
By selecting a smooth covariance function $k(\bm{\xi},\bm{\xi}')$ we can model
smooth functions. Furthermore, the kernel determines almost all the properties
of the resulting GP. For example, we can produce periodic processes by using a
periodic kernel. Since we are interested in non-parametric regression we shall
avoid those kernels that impose any predetermined form on the 
final predictor, e.g., linear kernels or polynomial kernels. When using flexible
kernels, GPs do not make strong assumptions about the nature of the function 
and, hence, they build their estimates from information derived from the data. 
Furthermore, even when lots of observations are used, there may still be some 
flexibility in the estimates. Thus, GPs are regarded as non-parametric methods
\cite[Chapter~1]{rasmussen2006gaussian}. Also, it should be noted that the 
number of parameters of a GP model grows with the amount of 
training data, which is another feature of non-parametric methods 
\cite[Section~1.4.1]{murphy2012machine}.

In the SDE estimation problem, we shall use two different GPs for modelling our 
prior beliefs about the properties of the 
drift and diffusion terms. After observing the data $\bm{x}$, we shall 
update our knowledge about them. This updated knowledge is represented by the 
posterior distributions  $p(\bm{f}^*|\bm{x})$ and  $p(\bm{s}^*|\bm{x})$, where 
$\bm{f}^*$ and $\bm{s}^*$ represent the sets that result from evaluating $f(x)$ 
and $g(x)$ over a set of inputs, i.e., $\bm{f}^* = \{f(x) : x \in \bm{x}^* \}$ 
(a similar expression applies to $g(x)$).

For the drift function $f$, we shall use a GP with zero mean. The zero mean 
arises from symmetry considerations and our lack of prior knowledge about $f$ 
(there is no reason to assume positive values instead of negative ones, and 
viceversa). On the other hand, we must ensure that $g > 0$ (since it plays the
role of a variance in Eq.~\eqref{eq:complete_lk}). Thus,  we assume that 
$g(x) = \exp \big(s(x)\big)$, where $s(x)$ is a Gaussian process with a constant 
mean function $m(x) = v$. The $v$ parameter is useful to control the scale of 
the noise process and possible numerical issues arising from the explosiveness 
of the exponential transformation. We shall use general covariance functions 
$\mathcal{K}_f$ and $\mathcal{K}_s$ for both processes, parametrized with the
hyperparameters $\bm{\theta}_f$ and $\bm{\theta}_s$, respectively. Hence, our 
complete model is:
\begin{subequations}
\label{eq:full_model}
\begin{align}
\begin{split}
\log p(\bm{x}|f, s, v) \approx {}& - \frac{1}{2}\sum_{i=1}^{N}i
\left[\frac{(\Delta x_i - f_i\Delta t)^2}{\exp \big(s_i\big)\Delta t} 
+ s_i\right] \\
&-\frac{N}{2}\log(2\pi\Delta t),
\label{eq:approx_lk}
\end{split}\\
f(x)|\bm{\theta}_f \sim {}& \mathcal{GP}
\big(0, \mathcal{K}_f(x, x',\bm{\theta}_f)\big),\\
s(x)|\bm{\theta}_s \sim {}&\mathcal{GP}
\big(v, \mathcal{K}_s(x, x',\bm{\theta}_s)\big),
\end{align}
\end{subequations}
where we have ignored the distribution of $p(x_1)$ from 
Eq.~\eqref{eq:complete_lk} (this is reasonable when $N >> 1$). Since $f(x)$ and 
$g(x)$ are GPs, the discrete vectors $\bm{f}=(f_1, f_2, \dots, f_N)$ and 
$\bm{s}=(s_1, s_2, \dots, s_N)$ must follow multivariate Gaussian distributions:
\begin{equation}
\label{eq:discrete_distros}
\bm{f}\mid \bm{\theta}_f \sim \mathcal{N}(\bm{0}^N, \bm{K}_{NN})\qquad 
\bm{s}\mid \bm{\theta}_s \sim \mathcal{N}(\bm{v}^N, \bm{J}_{NN}),
\end{equation}
where the entries of the covariance matrices are defined using
\begin{equation}
\left[\bm{K}_{NN}\right]_{ij} = \mathcal{K}_f(x_i, x_j, \bm{\theta}_f), \qquad
\left[\bm{J}_{NN}\right]_{ij} = \mathcal{K}_s(x_i, x_j, \bm{\theta}_s),
\label{eq:matrix_computation}
\end{equation}
and where $\bm{0}^N$ and $\bm{v}^N$ denote vectors of length $N$ with all their
entries set to $0$ and $v$, respectively.

Although in Eqs.~\eqref{eq:full_model}-\eqref{eq:matrix_computation} we have 
explicitly written  the dependencies on the hyperparameters 
$(v, \bm{\theta}_f\text{ and }\bm{\theta}_s)$ for the sake of completeness, we
shall assume, for the moment, that they are known and fixed. Hence, we shall
remove them from the equations in the next sections to keep the notation 
uncluttered.

As stated before, our aim is to compute the posteriors of any new set of new
function  points $\bm{f}^*$ (from $f(x)$) and $\bm{s}^*$ (from $s(x)$): 
$p(\bm{f}^*|\bm{x})$ and $p(\bm{s}^*|\bm{x})$. However, computing the posterior 
distribution of a model that involves GPs requires the calculation of inverse 
matrices, which usually scales as $\mathcal{O}(N ^ 3)$ operations 
\cite{rasmussen2006gaussian}. For large $N$, such as those used in the 
SDEs' literature, this 
approach is prohibitive. In the next section we discuss how to approximate the 
GP problem using only  $m$ points ($m << N$), which yields the so-called
Sparse Gaussian Processes (SGP)
\cite{rasmussen2006gaussian,quinonero2005unifying}.

\section{Approximation with Sparse Gaussian Processes \label{sec:sparse_gp}}
To overcome the intractable computations that a large dataset requires, many 
sparse methods construct an approximation to the GP using a small set of $m$
inducing variables ($m << N$).  Our inducing variables shall be the 
function points that result from evaluating $f(x)$ and $s(x)$ at some
pseudo-inputs $\bm{x}_m \in \mathbb{R}^m$, i.e. 
$\bm{f}_m = \{f(x): x \in \bm{x}_m \}$ and 
$\bm{s}_m = \{s(x): x \in \bm{x}_m \}$. Note that, although we could have used a
set of pseudo-inputs for $f(x)$ and another one for $s(x)$, we have opted for a
single set for the sake of simplicity. The key idea is that, instead of using 
the N-dimensional posterior distribution $p(\bm{f}\mid \bm{x})$ to compute the 
predictions of the new function points $\bm{f}^ *$, we could ``summarise" the 
information that we may learn from the data about $f(x)$ in a m-dimensional 
distribution $\phi_{f_m}(\bm{f}_m)$, and then use it to make the predictions 
(a similar reasoning also applies to $s(x)$). Since $\bm{f}_m$ and $\bm{s}_m$ 
represent the ``reference points" that we shall use to make new predictions, it
seems reasonable that $\bm{x}_m$ should be spread across the range
of values of $\bm{x}$.
Hence, in our problem, we shall require any pseudo-input to be contained in
$[\min \bm{x}, \max \bm{x}]$. It must be noted that the 
pseudo-inputs $\bm{x}_m$  may be seen as hyperparameters
of the complete model subject to optimization. However, for the moment we shall
assume that they are known and fixed. To determine how the pseudo-inputs can be 
used to make predictions, we shall  follow a similar approach to that introduced
in  \cite{titsias2009variational}, which used a variational formulation for 
learning the inducing variables of the SGP. The advantage of this approach is 
that variational inference naturally arises in our problem when trying to 
approximate the posterior distributions 
$p(\bm{f}|\bm{x})$ and $p(\bm{s}|\bm{x})$, as it will be discussed later.

Since $\bm{f}_m$
(or $\bm{s}_m$) and the $N$ points that result from evaluating $f(x)$ 
(or $s(x)$) at  $\{x_i\}_{i=1,2,...N}$ both sample the same GP, we assume: 
\begin{equation*}
   \begin{bmatrix}
     \bm{f}\\
     \bm{f}_m
   \end{bmatrix}
   \sim
  \mathcal{N}\left(
  \begin{bmatrix}
      \bm{0}\\
       \bm{0}
     \end{bmatrix}
  ,
  \begin{bmatrix} 
    \bm{K}_{NN} &\bm{K}_{Nm}\\
    \bm{K}_{mN} &\bm{K}_{mm}\\
  \end{bmatrix}\right),
\end{equation*}
\begin{equation*}
   \begin{bmatrix}
     \bm{s}\\
     \bm{s}_m
   \end{bmatrix}
   \sim
  \mathcal{N}\left(
  \begin{bmatrix}
      \bm{v}^N\\
       \bm{v}^m
     \end{bmatrix}
  ,
  \begin{bmatrix} 
    \bm{J}_{NN} &\bm{J}_{Nm}\\
    \bm{J}_{mN} &\bm{J}_{mm}\\
  \end{bmatrix}\right),
\end{equation*}
where all the submatrices are computed analogously as done in 
Eq.~\eqref{eq:matrix_computation}.

Since they will be used throughout the document, it is useful to write the 
expressions for the conditional distributions as:
\begin{equation*}
\begin{split}
\bm{f}|\bm{f}_m &\sim \mathcal{N}(\bm{A}\bm{f}_m, \bm{P}),\\
\bm{s}|\bm{s}_m &\sim \mathcal{N}(\bm{v}^N + \bm{B}(\bm{s}_m - \bm{v}^m), 
\bm{Q}),
\end{split}
\end{equation*}
with
\begin{equation}
\begin{split}
\bm{A} = \bm{K}_{Nm}\bm{K}_{mm}^{-1},&\qquad \bm{P} = \bm{K}_{NN} - 
\bm{K}_{Nm}\bm{K}_{mm}^{-1}\bm{K}_{mN},\\
\bm{B} = \bm{J}_{Nm}\bm{J}_{mm}^{-1},&\qquad \bm{Q} = \bm{J}_{NN} - 
\bm{J}_{Nm}\bm{J}_{mm}^{-1}\bm{J}_{mN}.
\label{eq:aux_matrices}
\end{split}
\end{equation}

Using this augmented model, the posterior distribution of the new function
points $\bm{f^*}$, from $f(x)$, and $\bm{s^*}$, from $s(x)$, would be: 
\begin{equation}
\begin{split}
p(\bm{f^*}, \bm{s^*} | \bm{x}) =\int & p(\bm{f^*},
\bm{s^*}| \bm{f}, \bm{f}_m, \bm{s},\bm{s}_m)\\
&\times p(\bm{f},\bm{f}_m, \bm{s},\bm{s}_m| 
\bm{x}) d\bm{f}d\bm{f}_md\bm{s}d\bm{s}_m\\
=\int & p(\bm{f^*}| \bm{f}, \bm{f}_m) p(\bm{s^*}| \bm{s}, \bm{s}_m)\\
& \times p(\bm{f},\bm{f}_m,\bm{s},\bm{s}_m| 
\bm{x}) d\bm{f}d\bm{f}_md\bm{s}d\bm{s}_m,
\end{split}
\label{eq:intractable_prediction}
\end{equation}

Following \cite{titsias2009variational}, we assume that $\bm{f}_m$ provides 
complete information for $\bm{f}^*$  in the sense that 
$p(\bm{f^*}|\bm{f}_m,\textbf{f}) = p(\bm{f^*}|\bm{f}_m)$. Similarly,
we assume $p(\bm{s^*}|\bm{s}_m,\textbf{s}) = p(\bm{s^*}|\bm{s}_m)$.
However, these assumptions do not prevent the computation of the GPs' posterior
$p(\bm{f},\bm{f}_m,\bm{s},\bm{s}_m| \bm{x})$. To make the model computationally 
efficient we shall approximate this distribution by factorizing
it in groups of $(\bm{f}, \bm{f}_m)$ and $(\bm{s}, \bm{s}_m)$, as
it is usually done in the variational inference approach 
\cite{titsias2009variational,BishopPR}:
\begin{equation}
\label{eq:approx_distros}
\begin{split}
p(\bm{f},\bm{f}_m\bm{s},\bm{s}_m | \bm{x}) &\approx 
\phi(\bm{f}, \bm{f}_m, \bm{s}, \bm{s}_m)\\
&=  p(\bm{f}|\bm{f}_m) 
\phi_{f_m}(\bm{f}_m) p(\bm{s}|\bm{s}_m)\phi_{s_m}(\bm{s}_m),
\end{split}
\end{equation}
where $\phi_{f_m}(\bm{f}_m)$ and $\phi_{s_m}(\bm{s}_m)$ denote unconstrained 
variational distributions over $\bm{f}_m$ and $\bm{s}_m$. Under this assumption, 
Eq.~\eqref{eq:intractable_prediction} becomes:
\begin{equation*}
\begin{split}
p(\bm{f^*}\mid \bm{x}) &\approx \int p(\bm{f^*}\mid\bm{f}_m)
\phi_{f_m}(\bm{f}_m) d\bm{f}_m,\\
p(\bm{s^*}\mid \bm{x}) &\approx \int p(\bm{s^*}\mid\bm{s}_m)
\phi_{s_m}(\bm{s}_m) d\bm{s}_m.\\
\end{split}
\end{equation*}

Given Eq.~\eqref{eq:approx_distros},  we may try to minimize the following 
Kullback-Leibler divergence to calculate the 
$\phi(\bm{f},\bm{f}_m,\bm{s},\bm{s}_m)$ distribution: 
\begin{equation}
\begin{split}
\mathcal{KL}(\phi\,|\, p) = \int & \phi(\bm{f}, \bm{f}_m, \bm{s}, \bm{s}_m)\\
&\times \log \frac{\phi(\bm{f}, \bm{f}_m,  \bm{s}, 
\bm{s}_m)}{p(\bm{f},\bm{f}_m,\bm{s},
\bm{s}_m|\bm{x})}d\bm{f}d\bm{f}_md\bm{s}d\bm{s}_m.
\end{split}
\end{equation}

Taking into account the identity  
\begin{equation*}
\log p(\bm{x}) = \mathcal{L}(\phi) + \mathcal{KL}(\phi\,|\, p),
\end{equation*}
where we have defined 
\begin{equation*}
\begin{split}
\mathcal{L}(\phi) = \int &  \phi(\bm{f},\bm{f}_m,\bm{s},\bm{s}_m)\\
&\times\log \frac{p(\bm{x}, \bm{f},\bm{f}_m,\bm{s},\bm{s}_m)}{
\phi(\bm{f},\bm{f}_m,\bm{s},\bm{s}_m)} 
d\bm{f}d\bm{f}_md\bm{s}d\bm{s}_m,
\end{split}
\end{equation*}
we notice that minimizing the Kullback-Leibler divergence with respect to 
$\phi$ is equivalent to maximize the lower bound of the marginal log-likelihood
$\mathcal{L(\phi)}$. Setting $\partial \mathcal{L(\phi)}/ \partial \phi = 0$ 
we obtain the optimal solutions \cite{BishopPR}:
\begin{subequations}
\label{eq:vi_sols}
\begin{equation}
\log \phi_{f_m}(\bm{f}_m) = \mathbb{E}_{\phi_{\uminus f_m}}\left[ 
\log\left(p(\bm{x}|\bm{f}, \bm{s})p(\bm{s}_m)p(\bm{f}_m)\right)\right],
\label{eq:vi_sol_f}
\end{equation}
\begin{equation}
\log \phi_{s_m}(\bm{s}_m) = \mathbb{E}_{\phi_{\uminus s_m}}\left[ 
\log\left(p(\bm{x}|\bm{f}, \bm{s})p(\bm{s}_m)p(\bm{f}_m)\right)\right],
\label{eq:vi_sol_s}
\end{equation}
\end{subequations}
where we have denoted: 
\begin{equation*}
\begin{split}
\phi_{\uminus f_m}(\bm{f}, \bm{s}, \bm{s}_m) &= 
p(\bm{f}|\bm{f}_m)p(\bm{s}|\bm{s}_m)\phi_{s_m}(\bm{s}_m)\text{, }\\
\phi_{\uminus s_m}(\bm{f}, \bm{f}_m, \bm{s}) &= 
p(\bm{f}|\bm{f}_m)\phi_{f_m}(\bm{f}_m)p(\bm{s}|\bm{s}_m),
\end{split}
\end{equation*}
to the density functions that result from ignoring the distributions 
$\phi_{f_m}(\bm{f}_m)$ and $\phi_{s_m}(\bm{s}_m)$ from 
$\phi(\bm{f}, \bm{f}_m, \bm{s}, \bm{s}_m)$ (see Eq.~\eqref{eq:approx_distros}),
respectively.

Note that Eqs.~\eqref{eq:vi_sols} are not a closed-form 
solution of the variational inference problem, since both equations are coupled. 
However, they naturally suggest the use of a coordinate ascent algorithm to 
find a solution. The coordinate ascent method iterates between holding
$\phi_{f_m}(\bm{f}_m)$ to update $\phi_{s_m}(\bm{s}_m)$ using 
Eq.~\eqref{eq:vi_sol_f} and holding $\phi_{s_m}(\bm{s}_m)$ to update 
$\phi_{f_m}(\bm{f}_m)$ through Eq.~\eqref{eq:vi_sol_s}. 

In our problem, Eq.~\eqref{eq:vi_sol_f} becomes (see Appendix \ref{ap:a}):
\begin{equation}
\begin{split}
\log \phi_{f_m}(\bm{f}_m) =& -\frac{1}{2}\bm{f}_m^{T}\left[ \bm{K}_{mm}^{-1} + 
\Delta t \bm{A}^{T} \text{diag}(\bm{\zeta})\bm{A}\right]\bm{f}_m \\
&+  \left[\bm{\zeta}\odot\bm{\Delta x}\right]^{T}\bm{A}\bm{f}_m + 
\text{constant},
\label{eq:f_sol}
\end{split}
\end{equation}
where $\odot$ denotes the element-by-element multiplication of two vectors,
$\text{diag}(\bm{\zeta})$ is the diagonal matrix constructed using the
values of the vector $\bm{\zeta}$ as main diagonal and
\begin{equation}
\label{eq:zeta}
\zeta_i = \mathbb{E}_{\phi_{s_m}}\left[
\exp{\left(-[\bm{v}^N + \bm{B}(\bm{s}_m - \bm{v}^m)]_i + \frac{Q_{ii}}{2}
\right)}\right],
\end{equation}
where $\bm{Q}$ was defined in Eq.~\eqref{eq:aux_matrices}.

Eq.~\eqref{eq:f_sol} implies that $\phi_{f_m}(\bm{f}_m)$ follows a Gaussian
distribution, that we shall write as:
\begin{equation}
\begin{split}
&\phi_{f_m}(\bm{f}_m) = \mathcal{N}(\bm{\mu}_f, \bm{F}), \text{ where}\\
&\bm{F} = \left[\bm{K}_{mm}^{-1} + \Delta t \bm{A}^{T}\text{diag}(\bm{\zeta})
\bm{A}\right]^{-1},\\
&\bm{\mu}_f = \bm{F}\bm{A}^{T}(\bm{\zeta}\odot \bm{\Delta x}).
\label{eq:normal_f}
\end{split}
\end{equation}

On the other hand, Eq.~\eqref{eq:vi_sol_s} becomes (see Appendix \ref{ap:b}):
\begin{equation}
\begin{split}
\log&{\phi_{s_m}(\bm{s}_m)}=\\
&-\frac{1}{2\Delta t}\sum_{i=1}^{N}\psi_i  
\exp{\left(-[\bm{v}^N + \bm{B}(\bm{s}_m - \bm{v}^m)]_i + 
\frac{Q_{ii}}{2}\right)}\\
&- \frac{1}{2}(\bm{s}_m - \bm{v}^m)^{T}\bm{J}_{mm}^{-1}(\bm{s}_m - \bm{v}^m)\\
&- \frac{1}{2} \sum_{i=1}^{N}[\bm{B}(\bm{s}_m - \bm{v}^m)]_i  + \text{constant},
\label{eq:s_sol} 
\end{split}
\end{equation}
where 
\begin{equation}
\begin{split}
\psi_i=\mathbb{E}_{\phi_{f_m}}\left[ \right.
&\Delta x_i^2 - 2 \Delta t\Delta x_i [\bm{A}\bm{f}_m]_i\\
&\left. + (\Delta t)^2\left([\bm{A}\bm{f}_m]_i^2 + P_{ii}\right) \right].
\end{split}
\label{eq:psi}
\end{equation}
Unlike the distribution for $\phi_{f_m}(\bm{f}_m)$, we cannot identify the 
distribution that appears in Eq.~\eqref{eq:s_sol}. This is not surprising, since 
the updates in a variational inference problem are only available in 
closed-form when using conditionally conjugate distributions. As a consequence,
we are forced to use approximate variational inference. 

\section{Laplace Variational Inference for the Estimation of the Diffusion 
\label{sec:laplace_approx}} Laplace approximations use a Gaussian to approximate 
intractable density functions. In the context of variational inference, it has 
already been considered in \cite{wang2013variational}, in order to handle 
non-conjugate models. We shall use this approach to handle Eq.~\eqref{eq:s_sol}.
Let  $\bm{\hat{s}}_m$ be the maximum of the right hand side from 
Eq.~\eqref{eq:s_sol}, which may be found using numerical optimization 
techniques. In our implementation of the method, we have used the L-BFGS-B 
algorithm \cite{byrd1995limited}, although any other method could have been
used. A Taylor expansion around ${\bm{\hat{s}}_m}$ gives:
\begin{align*}
\log{\phi_{s_m}(\bm{s}_m)} \approx&  \frac{1}{2}(\bm{s}_m - 
\bm{\hat{s}}_m)^{T}
\bm{H}_{\log{\phi}}(\bm{\hat{s}}_m)(\bm{s}_m - \bm{\hat{s}}_m)\\
& + \text{constant}
\label{eq:s_sol_laplace}
\end{align*}
where $\bm{H}_{\log{\phi}}(\bm{\hat{s}}_m)$ is the Hessian matrix of 
${\log{\phi_{s_m}  (\bm{s}_m)}}$ evaluated at $\bm{\hat{s}}_m$. In our case:
\begin{equation*}
\begin{alignedat}{3}
[\bm{H}_{\log{\phi}}(\bm{\hat{s}}_m)]_{kq} =& -\frac{1}{2\Delta t}
\sum_{i=1}^{N}&&\psi_i \cdot \exp{\left(-v +\frac{ Q_{ii}}{2}\right)}
B_{ik}B_{iq}\\
& && \times \exp{\left(-[\bm{B}(\bm{\hat{s}}_m- \bm{v}^m)]_i\right)}\\
&-[\bm{J}_{mm}^{-1}]_{kq}.&&
\end{alignedat}
\end{equation*}

Thus, the approximate update for $\phi_{s_m}(\bm{s}_m)$ to be used in the
coordinate ascent algorithm is a Gaussian distribution:
\begin{equation}
\begin{split}
&\phi_{s_m}(\bm{s}_m) \approx \mathcal{N}(\bm{\mu}_s, \bm{S}), \text{ where}\\
&\bm{\mu}_s = \bm{\hat{s}}_m,\qquad \bm{S} = -[\bm{H}_{\log{\phi}}
(\bm{\hat{s}}_m)]^{-1}.
\end{split}
\label{eq:normal_s}
\end{equation}

Taking into account that both the distributions of $\bm{f}_m$ and 
$\bm{s}_m$ are Gaussians, we can write $\psi_i$ and $\zeta_i$ as:
\begin{equation}
\begin{split}
\zeta_i =& \exp{\left[-[v + \bm{B}(\bm{\mu}_s - \bm{v}^m)]_i + \frac{1}{2} 
(Q_{ii} + \bm{B}_{i,.}^{T}\bm{S}\bm{B}_{i,.})\right]},\\
\psi_i =& (\Delta x_i)^2 -2 \Delta t \Delta x_i [\bm{A}\bm{\mu}_f]_i\\
        &+(\Delta t) ^ 2 [\bm{P} + \bm{A} (\bm{\mu}_f \bm{\mu}_f^{T} +
        \bm{F})\bm{A}^{T}]_{ii}, 
\end{split}
\end{equation}
where $\bm{B}_{i,.}$ denotes the $i$-th row of the $\bm{B}$ matrix. It must be
noted that the values of $\bm{\psi}$ and $\bm{\zeta}$ should be updated with 
each step of the coordinate ascent algorithm. After each step, the convergence 
of the  algorithm must be assessed by computing the lower bound 
$\mathcal{L}(\phi)$:
\begin{equation}
\begin{split}
\mathcal{L}(\phi) =& \mathbb{E}_\phi\left[\log p(\bm{x}|f,s)\right]\\
&+\mathbb{E}_{\phi_{f_m}}\left[\log p(\bm{f}_m)\right] + 
\mathbb{E}_{\phi_{s_m}}\left[\log p(\bm{s}_m)\right]\\
&-\mathbb{E}_{\phi_{f_m}}\left[\log \phi_{f_m}(\bm{f}_m)\right] 
-\mathbb{E}_{\phi_{s_m}}\left[\log\phi_{s_m}(\bm{s}_m)\right]\\
=&-\frac{1}{2}\sum_{i=1}^{N}\mathbb{E}_\phi\left[\frac{(\Delta x_i - 
\Delta t f_i)^2}{\Delta t \exp{(s_i)}} - s_i\right] \\
&-\frac{N}{2} \log (2\pi \Delta t ) -\frac{1}{2} \log |\bm{K}_{mm}| 
-\frac{m}{2} \log 2\pi \\
&-\frac{1}{2}\mathbb{E}_{\phi_{f_m}}\left[\bm{f}_m^T
\bm{K}_{mm}^{-1}\bm{f}_m\right] \\
&-\frac{1}{2} \log |\bm{J}_{mm}| -\frac{m}{2} \log 2\pi \\
&-\frac{1}{2}\mathbb{E}_{\phi_{s_m}}\left[(\bm{s}_m - 
\bm{v}^m)^T\bm{J}_{mm}^{-1}(\bm{s}_m - \bm{v}^m)\right]\\
&+\mathbb{H}_{\phi_{f_m}}\left[\bm{f}_m\right] + \mathbb{H}_{\phi_{s_m}}
\left[\bm{s}_m\right],
\label{eq:lower_bound_expectation}
\end{split}
\end{equation}
where $\mathbb{H}$ is the entropy of a distribution. Taking
the expectations in Eq.~\eqref{eq:lower_bound_expectation} yields
\begin{equation}
\begin{split}
\mathcal{L}(\phi) =&-\frac{1}{2\Delta t}\sum_{i=1}^{N} \psi_i\zeta_i
-\frac{1}{2}\sum_{i=1}^{N} [\bm{v}^N + \bm{B}(\bm{\mu}_s - \bm{v}^m)]_i \\
&-\frac{N}{2}\log (2\pi \Delta t) -\frac{1}{2} \log |\bm{K}_{mm}|
-\frac{m}{2} \log 2\pi \\
&-\frac{1}{2} \left[\text{tr}(\bm{K}_{mm}^{-1}\bm{F}) 
+\bm{\mu}_f^T \bm{K}_{mm}^{-1} \bm{\mu}_f\right]\\
&-\frac{1}{2} \log |\bm{J}_{mm}| -\frac{m}{2} \log 2\pi \\
&-\frac{1}{2} \left[\text{tr}(\bm{J}_{mm}^{-1}\bm{S}) 
+(\bm{\mu}_s - \bm{v}^m)^T \bm{J}_{mm}^{-1} (\bm{\mu}_s - \bm{v}^m)\right] \\
&+ \frac{1}{2}\log{\big((2\pi e)^m|\bm{F}|\big)}
+ \frac{1}{2}\log{\big((2\pi e)^m|\bm{S}|\big)}\\
&+ \text{constant},
\end{split}
\label{eq:gaussian_lower_bound}
\end{equation}
where $\text{tr}(\cdot)$ denotes the trace of a matrix. In addition to checking 
the  convergence, computing the lower bound permits checking the correctness of
the  implementation since it should always increase monotonically; and since it
is an approximation to the marginal likelihood, it can be used for Bayesian 
model selection. For example, we can use the lower bound to select the best kernel 
among a set of possible ones or to select the number of inducing points $m$. 
However, there is a subtle detail that must be addressed. Although the 
variational inference  framework approximates the posterior distribution, it 
only does it around one of the local modes. With $m$ pseudo-inputs, there are 
$m!$ equivalent modes due to the lack of identifiability of the pseudo-inputs
(the different  modes only differ through a relabelling of 
the $\bm{x}_m$ vector). A simple approximate solution that takes into account 
the multi-modality is using: 
\begin{equation}
\mathcal{L'} \approx \mathcal{L} + \log(m!),
\label{eq:L_comparison}
\end{equation}
for model selection \cite{BishopPR}.

\section{Hyperparameter Optimization \label{sec:hp_optimization}}
So far, we have assumed that the ``variance" parameter $v$, the 
hyperparameters of the covariance functions, $\bm{\theta}_f$ and 
$\bm{\theta}_s$, and the  pseudo-inputs $\bm{x}_m$ were known and fixed. 
However, Eq.~\eqref{eq:gaussian_lower_bound} does depend on all these
hyperparameters, i.e.  $\mathcal{L}(\phi) \equiv \mathcal{L}(\phi,v, 
\bm{\theta}_f, \bm{\theta}_s, \bm{x}_m) = \mathcal{L}(\phi, \bm{\theta}_{\text{all}})$,
and hence, further maximization of the lower bound could be achieved. Note that
this  optimization permits the automatic selection of the
inducing-inputs $\bm{x}_m$ and the kernel hyperparameters starting from some
reasonable initial values. In our 
implementation, we have interleaved the updates of the variational 
distributions with the numerical optimization of 
the lower bound with respect to the hyperparameters 
(since the analytical optimization is intractable). This permits the slow 
adaptation of the hyperparameters to the variational distributions. The 
resulting algorithm may be compared with a Generalized Expectation Maximization
algorithm (GEM) \cite{mclachlan2007algorithm}. In what we may identify as the E 
step, the variational distributions are updated. First, the distribution
parameters $\bm{\mu}_f$ and $\bm{F}$ are modified according to 
Eq.~\eqref{eq:normal_f} using the last values obtained for $\bm{\mu}_s$ and 
$\bm{S}$ to compute any expectation involving the random variable $\bm{s}_m$.
These new values are then used to compute the expectations involving $\bm{f}_m$
and updating $\bm{\mu}_s$ and $\bm{S}$ through Eq.~\eqref{eq:normal_s}. 
In the M step, the lower bound given by Eq.~\eqref{eq:gaussian_lower_bound} is
further optimized with respect to the hyperparameters while keeping the
distribution parameters $(\bm{\mu}_f, \bm{F}, \bm{\mu}_s, \bm{S})$ fixed. 
Given that finding a maximum may have a slow convergence, instead of aiming to 
maximize the lower bound we sought to change the hyperparameters in such a way 
as to increase it: $\mathcal{L}(\phi, \bm{\theta}^{n+1}_{\text{all}}) >
\mathcal{L}(\phi, \bm{\theta}^{n}_{\text{all}})$. This may be interpreted as a
``partial'' M step. In  our implementation, we just limited the number of
iterations of a L-BFGS-B algorithm \cite{byrd1995limited}, although any other
numerical method could have been used. Changing from the maximization of the
objective to simply searching for an increase of it is what makes 
our method similar to the GEM algorithm instead of the standard EM algorithm. 
The E and M steps are then repeated until the convergence of the lower bound
$\mathcal{L}$.

\subsection{Hyperparameter Initialization and Kernel Selection\label{sec:hp_initialization}}
The lower bound in a variational problem is usually a non-convex function and 
hence, the proposed GEM-like algorithm is only guaranteed to converge to a local
maximum, which can be sensitive to initialization \cite{blei2017variational}. 
Thus, several trials with randomly selected initial values of the
hyperparameters should be run. The final estimate can be selected using 
Eq.~\eqref{eq:L_comparison}. However, it should be noted that, 
experimentally, solutions stacked in a clearly suboptimal local maxima happen 
infrequently.

Given that SGPs provide a Bayesian framework, the kernels and the initial values
for their hyperparameters should be selected to model the prior beliefs about 
the behaviour of the drift and diffusion functions. Choosing a proper kernel
requires some knowledge about the properties of covariance functions
\cite[Chapter~4]{rasmussen2006gaussian} and 
experience to combine them to model functions with different kinds of structure 
\cite[Chapter~2]{duvenaud2014automatic}. Reasonable choices 
commonly used in the GP literature when no
prior information is available are the squared exponential kernel (or 
Gaussian kernel) and the rational quadratic kernel 
\cite{rasmussen2006gaussian}, although any kernel 
could be used within our method. The squared exponential kernel is one of the 
most widely used  covariance functions in the field of GP regression since it
is infinitely differentiable and hence it yields very smooth processes 
\cite[Chapter~2]{rasmussen2006gaussian}. Its main hyperparameter is the 
length-scale $l$, i.e., the variation necessary in the input variable for the
function values to appreciably change. On the other hand, the rational quadratic
kernel can be seen as an infinite sum of squared exponential covariance 
functions with different length-scales. It has two main hyperparameters, a mean
length-scale $l$ and a parameter controlling the mixing of the different 
squared exponential kernels (derived from a gamma distribution) 
\cite[Chapter~4]{rasmussen2006gaussian}. 

The amplitude of a kernel function can be interpreted as the prior belief about 
the variance of the drift/diffusion term. Hence, large amplitudes can be used
when no prior information is available. The selection of the amplitude
hyperparameter for the diffusion requires further discussion since we have to 
link the amplitude of the kernel modelling $s(x)$, $A_s$, with our prior belief 
about the variance of $g(x) = \exp \big(s(x)\big)$, $A_g$. Furthermore, it also
requires selecting an initial value for $v$. Since a lognormal random variable 
$Z \sim \log \mathcal{N}(\mu=v, \sigma ^2=A_s)$ fulfils:
\begin{equation}
\mathbb{E}\left[Z\right] = e^{v + \frac{A_s}{2}}, \qquad 
\text{Var}\left[ Z \right] = (e^{A_s} - 1)e^{(2v + A_s)},
\label{eq:lognormal}
\end{equation}
we find the proper parameters $v$ and $A_s$ from the prior belief $A_g$ and the
data itself, $\bm{x}$, using:
\begin{equation}
\begin{split}
A_s &= \log \bigg( 1  + \frac{A_g}{ (\text{Var}\left[ \Delta \bm{x}\right]/ 
\Delta t)^2 }\bigg),\\
v &= \log \bigg( \frac{\text{Var}\left[\Delta \bm{x} \right]}{ \Delta t } \bigg) - 
\frac{A_s}{2}.
\end{split}
\label{eq:v_selection}
\end{equation}

As argued in Section \ref{sec:sparse_gp}, $\bm{f}_m$ and $\bm{s}_m$ may be 
interpreted as ``reference points" used to infer the shape of $f(x)$ and 
$s(x)$. Hence, we may expect $\bm{x}_m$ to be spread across the range of values
of $\bm{x}$ so that the function shapes can be properly modelled in the whole 
range of $x$. It is also reasonable to assume that the inducing points should be
more concentrated in those regions where $f(x)$ or $s(x)$ change their curvature.
However, in our non-parametric approach, we cannot presume any prior knowledge
about these regions. Thus, a simply strategy for selecting the initial 
values of the pseudo-inputs would be to uniformly spread $\bm{x}_m$ between
$\min \bm{x}$ and $\max \bm{x}$. It is possible to design another 
approach based on the inducing points tending to regions with
low uncertainty about the function shape. This is due to the fact that the 
inducing points permit reducing the variance around their 
``region of influence'', which enables accurately modelling the 
low-uncertainty true posterior and hence reducing the Kullback-Leibler
divergence. Further evidence about this will given in Section 
\ref{sec:synthetic}. Thus, we propose
to initialize the inducing points to the result of applying the 
quantile function to the values $\{0/(m-1), 1/(m-1), ..., (m-1)/(m-1)\}$, since
this approach concentrates the inducing points in the region where more evidence
for inferring confident estimates is available. We will later refer to this
approach as the ``percentile initialization''. It must be noted that, when 
performing several runs of the estimation algorithm, random noise can be added 
to each value of $\bm{x}_m$ to obtain slightly different starting points. 
In practical applications, we also add the
restriction that, after adding the noise, the $\bm{x}_m$ vector should remain 
ordered and that $\min{\bm{x}_m} \geq \min\bm{x}$ and $\max{\bm{x}_m} 
\leq \max\bm{x}$.

The selection of the number of inducing points $m$ is the most challenging one
since SGP usually get better approximations to the full GP posterior  when using
more points (larger $\mathcal{L'}$), at the cost of greater computational time 
\cite{rasmussen2006gaussian}. When taking into account both factors, there is 
not an unique way of defining which is the optimum value of $m$ and hence, the
final choice can be subjective. Rasmussen et al. suggest to perform runs with
small  values of $m$ and compare the resulting estimates between them while
getting a feeling on how the running time scales \cite{rasmussen2006gaussian}.
Since most kernels use a length-scale parameter $l$ we suggest using
\begin{equation}
m = \lfloor (\max \bm{x} - \min \bm{x}) / l \rfloor
\label{eq:m_heuristic}
\end{equation}
as a rule of thumb for getting an estimate of a proper number of inducing 
points. This rule uses only a few inducing points when the function varies 
very smoothly (large $l$) and a large number of them when the function  
wiggles quickly (small $l$). 

\section{Validation on synthetic data \label{sec:synthetic}}
To assess the validity of the SGP method, we compare its performance
with the kernel based method \cite{lamouroux2009kernel} and with a version
of the orthonormal polynomials method  \cite{rajabzadeh2016robust} using a set
of simulated SDEs.  From now on, we shall refer to these methods as the 
KBR (Kernel Based Regression) and the  POLY method 
(since it is based on  orthonormal polynomials), respectively. We have included 
the POLY method since it is described as non-parametric in 
\cite{rajabzadeh2016robust}, although we find it closer to a parametric one 
(see Section \ref{sec:intro}). We have also used 
these tests to further investigate the impact of the number of pseudo-inputs $m$ 
on the estimates.

For the validation, we consider the generic SDE described by Eq.~\eqref{eq:sde}
parametrized with the drift and diffusion functions summarized
in Table \ref{tab:models}.  It must be noted that some of these tests have 
been inspired by some well-known models. $M_1$ is the celebrated
Ornstein-Uhlenbeck model, which describes the motion of a Brownian particle
in velocity space \cite{langevin1908theorie}. $M_4$ is the Jacobi diffusion
process, which has an invariant distribution that is uniform on $(0,1)$ 
\cite{iacus2009simulation}. A Jacobi based model was used in 
\cite{larsen2007diffusion} to model exchange rates in target zone. $M_5$ is the
Cox-Ingersoll-Ross model. Despite it was introduced to model population growth,
it has become popular after its proposal for studying short-term interest rates 
in finance \cite{cox1985theory}. Although $M_2$ and $M_6$ do not receive any 
particular name, they are interesting models since they are able to generate
time series with a bimodal density. Finally, $M_3$ was used to test dynamical 
systems with nonlinear drift and diffusion functions and just a single stable
point.

For each of these models, 100 time series with a 
length of $10^4$ samples were generated. The Euler-Maruyama scheme
with an integration step  $\Delta t = 0.001$ was used for the simulations. The 
quality of the estimations obtained for the $i$-th simulation of Model
$M_j$ was assessed by the weighted integrated absolute error:
\begin{equation}
\mathcal{E}(M_j, i) = \int_{-\infty}^\infty \lvert F(x) - 
\hat{F}(x) \rvert \cdot p_i(x) dx,
\label{eq:integrated_error}
\end{equation}
where $F$ can be either $f$ or $g$, $\hat{F}$ denotes its estimate and $p_i(x)$
is the probability density function of the $i$-th simulation of the $M_j$
model. In practice, $p_i(x)$ is approximated using a kernel density estimate 
with a Gaussian kernel. The bandwidth of the kernel is selected using 
Silverman's ``rule of thumb" 
\cite[Page~48, Equation~3.31]{silverman1986density}.

To select a proper bandwidth for the KBR method, the 
selection algorithm described in \cite{lamouroux2009kernel} was implemented.
Regarding the POLY method, the parameter estimation was performed with
polynomials of orders $R = 1,2,\dots, 5$ and $L = 0,1, \dots ,3$ for the drift
and the  diffusion terms, respectively. Instead of using the Legendre 
polynomials as in  \cite{rajabzadeh2016robust}, the orthonormal polynomials
described  in \cite{kennedy1980statistical} were employed for easiness of 
implementation. Our tests indicate that the use of these polynomials instead of 
the Legendre polynomials do not undermine the expressive power of the method. 
Three  different  model selection methods were tested 
within the POLY framework. The  simulation based method proposed in 
\cite{rajabzadeh2016robust}, a  cross-validation method and a stepwise  
regression method. Since the later yielded the best results, we shall focus on 
it. The stepwise regression method that we have implemented uses a bidirectional
elimination approach. It starts with no predictors for the drift function. 
Then, at each step until convergence, it adds or removes an orthonormal 
polynomial term by  comparing the AIC (Akaike Information Criterion) 
improvement that results from each possible decision. The procedure stops when
no more predictors can be added or removed from the model. The method is then 
repeated for the diffusion term.

Regarding the SGP method, the same kernel was selected for estimating both the
drift and diffusion terms: 
\begin{equation}
\begin{split}
&\mathcal{K}(\bm{\xi} , \bm{\xi}', A, \bm{\theta}) =  \theta_0 
\exp \left[-\frac{\theta_1}{2}
\|\bm{\xi} - \bm{\xi}'\| ^ 2\right] + (A - \theta_0). 
\label{eq:general_kernel}
\end{split}
\end{equation}
The kernel $\mathcal{K}$ is a linear combination of a squared 
exponential kernel (first term in the right-hand side  (RHS)) and a constant
kernel (second term in the RHS). Note that the hyperparameter $\theta_1$ 
determines the characteristic length-scale of the GP ($l^2=1/\theta_1$). 
The constant covariance function was included since a constant diffusion term is
often used in the literature. It must be noted that we have not treated the 
parameter $A$ as an hyperparameter subject to optimization (we have not included 
it into the hyperparameter vector $\bm{\theta}$). We prefer to keep it fixed so 
that the total amplitude of the diagonal of the covariance matrices that result 
from $\mathcal{K}$ always sum up to $A$. In this way, $A$ can be interpreted as
the prior belief about the variance of the drift/diffusion term. This eases the
comparison between several optimization runs using  Eq.~\eqref{eq:L_comparison},
since all the estimates share the same prior belief about the range in which the
dynamic terms may lie. Note that in order to fulfil 
$\mathcal{K}(\bm{\xi} , \bm{\xi}', A, \bm{\theta}) \in [0, A]$ we must 
perform a box-constrained optimization of $\theta_0$ ($\theta_0 \in [0,A]$),
which originally  motivated the use of the L-BFGS-B  method as the 
optimization algorithm.

Since it is usual to get ill-conditioned  covariance matrices when working with
GPs we slightly modified Eq.~\eqref{eq:general_kernel}. To regularise the 
covariance matrices a small value on  the principal diagonals was added. In 
general, any type of kernel $\mathcal{Q}$ can be modified to improve stability 
as:
\begin{equation}
\mathcal{Q}'(\bm{x}, \bm{x}', \bm{\theta}, \epsilon) = \mathcal{Q} (\bm{x}, 
\bm{x}', \bm{\theta}) + \epsilon \delta(\bm{x} - \bm{x}'),
\label{eq:modified_kernel}
\end{equation}
where we did not state the dependencies of $\mathcal{Q}$ that are not treated as 
hyperparameters (e.g., $A$ in Eq.~\eqref{eq:general_kernel}).
When using the modified squared exponential kernel $\mathcal{K}'$, we did not
optimize on the $\epsilon$ parameter to
avoid creating large discontinuities in the covariance function.

\begin{figure}
\includegraphics[width=3.38in]{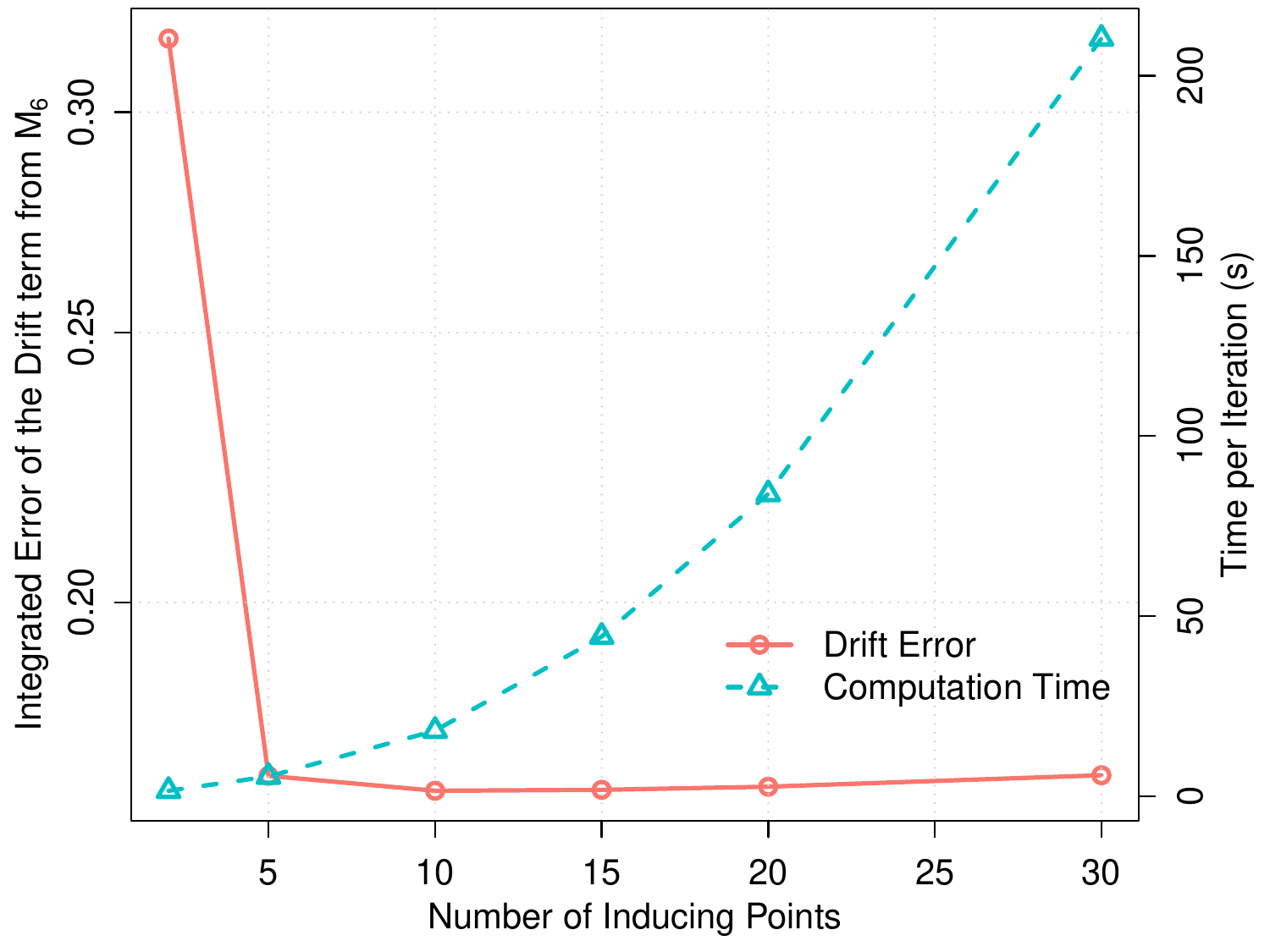}
\caption{Drift's integrated error and computational time per iteration depending on $m$
for a small subset of $M_6$ simulations.}\label{fig:m_vs_time}
\end{figure}

Since all the models used for testing have very smooth functions and they generate 
time series with a range of the order of 1, we may expect good estimates with 
only a few inducing-points. For example, the drift function of $M_2$ has three roots
at $-1$, $0$ and $1$ and, therefore, a reasonable estimate for its length-scale
would belong to $[0.5, 1]$. A typical trajectory of $M_2$ would probably lie in the interval
$x(t) \in [-2, 2]$ and hence, an estimate of the $m$ based on 
Eq.~\eqref{eq:general_kernel} would yield $m = 4 / 0.5 = 8$. To verify our intuitions,
we have followed Rasmussen's approach \cite{rasmussen2006gaussian} (see 
Section~\ref{sec:hp_initialization}). We have calculated the integrated error
of the drift function for $M_6$ on a small
subset of simulations while testing how the computation time scales with $m$. 
The drift function for $M_6$ was selected for the test since it is probably
the most complex one.
Fig.~\ref{fig:m_vs_time} shows that there are not big differences in the 
integrated errors for $m \geq 10$, whereas the time per iteration 
quickly scales.

Based on these results, we run our SGP method using $m=2,5,10$ and
$15$. It should be noted that $m << N$ and hence, we could have used larger $m$ without 
compromising the computational tractability of the problem. Also note that
although Fig.~\ref{fig:m_vs_time} suggests
that we could stop searching at $m=10$, we have also included $m=15$. This was done
to compare both estimates and further investigate the impact of $m$ in the
lower bound. Furthermore, Fig.~\ref{fig:m_vs_time} was
obtained using a small subset of the data from a single model and, 
therefore, there may be  simulations for which the use of $m=15$ may yield 
better estimates.

For each value of $m$,
several trials with randomly selected initial values of the hyperparameters were
run. The length-scales of both drift and diffusion kernels were  restricted to 
the interval $ l\in [0.25, 2]$, based again on the fact that all 
the time series have a range of the order of 1. The value $A_f$ was set to 
25 (equivalent to a standard deviation of 5) and the initial value of
$\theta_{f,0}$ was randomly initialized into the interval $[0, A_f]$. The 
selection of $v$ and the amplitude hyperparameter for the diffusion process
was made using Eq.~\eqref{eq:v_selection} and $A_g = 25$ for all 
the models present in the simulated set. The starting values for the 
pseudo-inputs were selected using the percentile initialization. The final
model for each of the time series was selected by using the modified lower bound
(Eq.~\eqref{eq:L_comparison}).

Table \ref{tab:errors} summarises the mean values of the integrated errors
for all the models from Table \ref{tab:models}. The best result for each model
is marked in bold (smaller is better) \footnote{For further reproducibility,
the parameters inferred in the algorithm can be found at 
\url{https://github.com/citiususc/voila/blob/master/additional_material/synthetic_data_parameters.txt}}. 
Additionally, a star (*) points those best-results with statistically
significant  differences with respect to the other two methods. The differences
between  methods were tested using the Nemenyi post-hoc test 
\cite{nemenyi1962distribution}. The results in Table \ref{tab:errors} show that
our proposal has a good performance, specially in the drift estimates, where it
performs better than KBR and POLY in the majority of the
models. The results for the diffusion are also good, but the SGP 
method has the largest mean error for the $M_5$ model. The reason for this
is discussed below.

\begin{table}
\caption{Models used for the validation with synthetic data.}
\label{tab:models}
\begin{ruledtabular}
\begin{tabular}{ccc}  
Model & $f(x)$ & $\sqrt{g(x)}$ \\ 
\hline
$M_1$ & $-(x - 3)$ &  $\sqrt{2}$ \\ 
$M_2$ & $-(x^3 -x)$ &  $1$\\ 
$M_3$ & $-x^3$ & $0.2 + x^2$\\ 
$M_4$ & $-0.7 (x - 0.5)$ & $\sqrt{0.7 x (1 - x)}$\\ 
$M_5$ & $-(x - 0.225)$ & $0.5 \sqrt{x}$\\
$M_6$ & $-x + \sin(3.5* x)\exp(-x^2)$ &  $0.431$\\ 
\end{tabular}
\end{ruledtabular}
\end{table}

\begin{table*}[ht]
\caption{Integrated absolute errors of the methods KBR and 
POLY and our proposal (denoted as SGP), using different test
models with length $N=10^4$.}
\label{tab:errors}
\begin{ruledtabular}
\begin{tabular}{c......}
&\multicolumn{3}{c}{Drift estimates} &\multicolumn{3}{c}{Diffusion Estimates}\\
\multicolumn{1}{c}{\textrm{Model}}& 
\multicolumn{1}{c}{\phantom{YYY}KBR}& 
\multicolumn{1}{c}{\phantom{YYY}POLY}& 
\multicolumn{1}{c}{\phantom{YYY}SGP}& 
\multicolumn{1}{c}{\phantom{YYY}KBR}& 
\multicolumn{1}{c}{\phantom{YYY}POLY}& 
\multicolumn{1}{c}{\phantom{YYY}SGP}\\
\hline
$M_1$ & 0.6863 & 0.7896  & \multicolumn{1}{Z{.}{.}{-1}}{0.4992*}
& 0.03963 &  0.03426  & \multicolumn{1}{Z{.}{.}{-1}}{0.02684*}\\
$M_2$ & \multicolumn{1}{Z{.}{.}{-1}}{0.5073*} & 0.6267 & 0.5760 &
0.01915  & 0.01878 & \multicolumn{1}{Z{.}{.}{-1}}{0.01511*}\\
$M_3$ & 0.1501  & 0.2731 & \multicolumn{1}{Z{.}{.}{-1}}{0.1232*} &
0.05293 & 0.02711 &\multicolumn{1}{Z{.}{.}{-1}}{0.007465*}\\
$M_4$ & 0.1244 & 0.1519 & \multicolumn{1}{Z{.}{.}{-1}}{0.1128*} &
0.02585 & \multicolumn{1}{Z{.}{.}{-1}}{0.002054*} & 0.0045\\
$M_5$ & 0.09035 & 0.1613 & \multicolumn{1}{Z{.}{.}{-1}}{0.08256*} &
\multicolumn{1}{Z{.}{.}{-1}}{0.001338*} & 0.001771 & 0.002667\\
$M_6$ & 0.2289 & 0.2618 & \multicolumn{1}{Z{.}{.}{-1}}{0.2256} & 
0.002751 & 0.002972 & \multicolumn{1}{Z{.}{.}{-1}}{0.002323*} \\
\end{tabular}
\end{ruledtabular}
\end{table*}

\begin{figure*}
\includegraphics[width=6.5in]{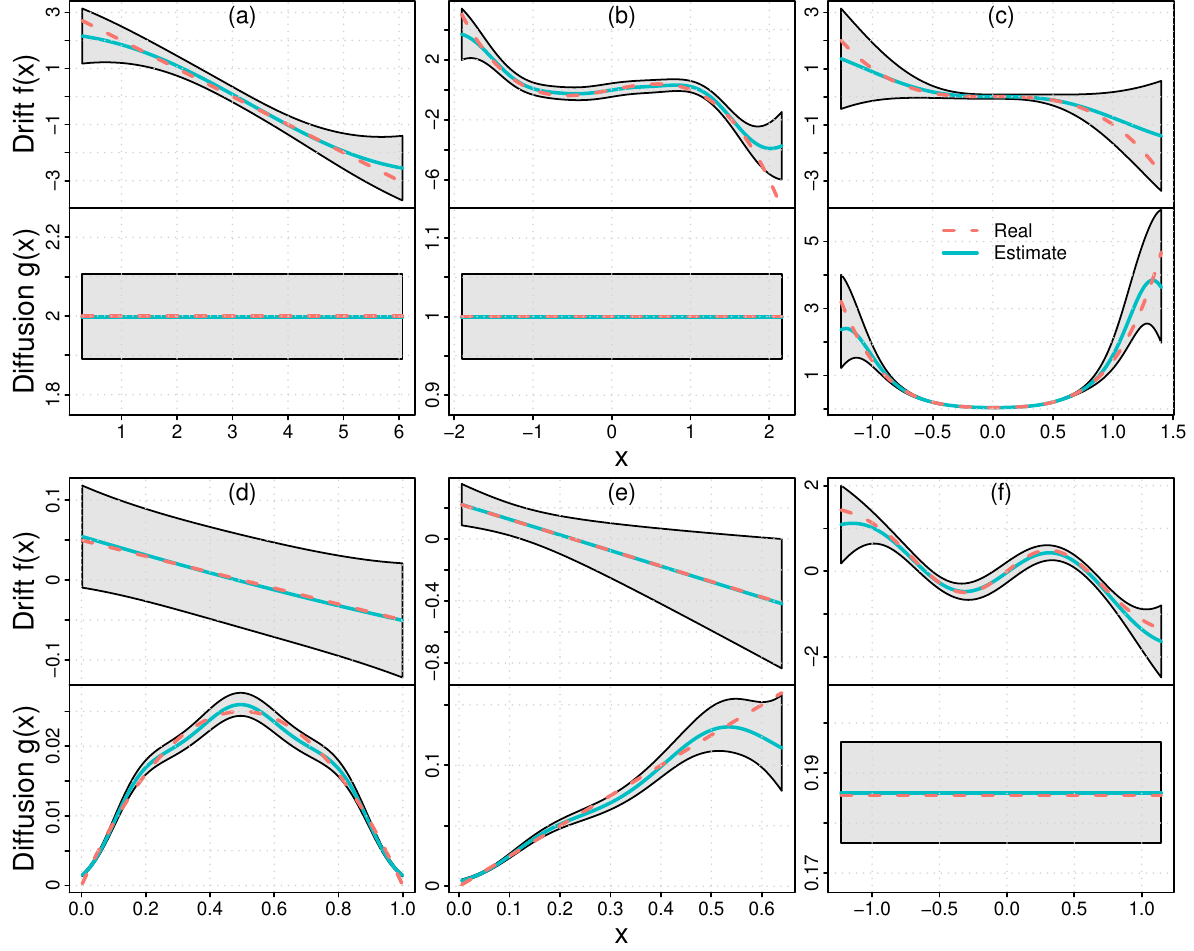}
\caption{Drift and diffusion estimates obtained from a 
single trajectory of the simulated models: (a) $M_1$, (b) $M_2$, (c) $M_3$,
(d) $M_4$, (e) $M_5$ and (f) $M_6$. The shaded area represents 
the 95\% confidence region.}\label{fig:estimate_examples}
\end{figure*}
   
Fig.~\ref{fig:estimate_examples} illustrates the kind of estimates that the SGP 
method yields for the drift and diffusion terms from a single 
realization of the simulated models. Note that the confidence intervals (grey
regions) usually increase when $x$ takes extreme values. This is due to the fact
that the regions where $x$ takes extreme values are only visited a few times
during any simulated trajectory and hence only a few points are available for
the estimation. Since there is little data at these regions, the priors have 
strong influence and the estimates tend to curve towards the prior means. This
effect is particularly remarkable for the drift estimates, which curve towards 
zero, and the diffusion for $M_5$. This is probably the reason why the 
SGP method does not perform as well as expected for the diffusion for $M_5$ and
the drift for $M_2$.

\begin{figure}
\includegraphics[width=3.38in]{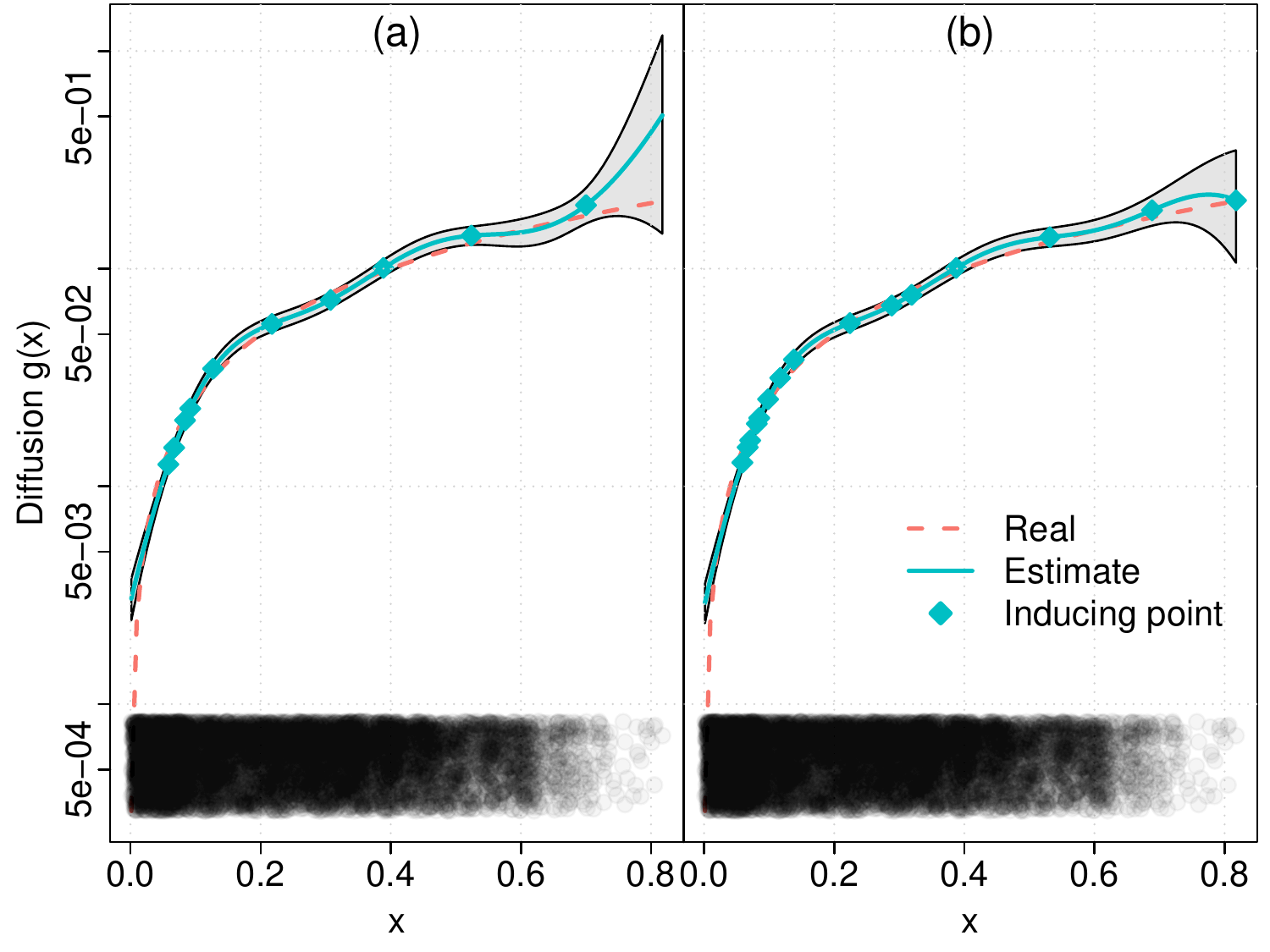} 
\caption[Influence of the inducing points on the estimates]{Diffusion estimates
obtained using (a) $m=10$ and (b) $m=15$ inducing points. The density of the
point cloud at the bottom of the figure represents the number of points
available for the estimation at each $x$.}\label{fig:why_large_m}
\end{figure}

Concerning the selected number of pseudo-inputs $m$, the general trend is 
that $\mathcal{L'}$ (see Eq.~\eqref{eq:L_comparison}) increases  with $m$, as we
might have expected (see Section \ref{sec:hp_initialization}). Hence, all
the selected models use $m=15$ inducing points. However, it is not always worth 
to increase $m$ in terms of the integrated error versus the running time, which
scales as $\mathcal{O}(2^m)$ due to the use of the L-BFGS-B algorithm (see
Fig.~\ref{fig:m_vs_time}). This can be understood by looking at 
Fig.~\ref{fig:why_large_m}. The figure shows two estimates  of the $M_5$'s 
diffusion term obtained using a different number of inducing-points, which are 
also represented in the plot. As noted with Fig.~\ref{fig:estimate_examples}, 
the width of the confidence intervals (grey regions), depends on  the number
of points available for the estimation, illustrated with the point cloud. The 
similarity between both estimates over the high-density region results in an 
almost identical weighted integrated error. However, the  $\mathcal{L'}$ is
larger for $m = 15$  than for $m = 10$, mostly because the
confidence interval significantly increases in the low-density region for 
$m = 10$. The use of additional inducing-points  in the case $m = 15$ permits a
better control of the estimates and the confidence interval, which results in a
larger $\mathcal{L'}$ although the weighted integrated error is very similar.
Hence, the $\mathcal{L'}$ based selection criteria is not optimal for the 
purpose of minimizing the weighted integrated error without wasting 
computational resources. From these experimental results about the
impact of $m$ in $\mathcal{L'}$ we conclude the selection of $m$ should
not be based solely on the lower bound, since it monotically increases with $m$
at a cost of greater computation times. Therefore, we suggest adopting
Rasmussen's heuristic (Section~\ref{sec:hp_initialization}) in combination
with $\mathcal{L'}$, using Eq.~\eqref{eq:m_heuristic} as an initial guess 
for the value of $m$.

\section{Application to real data \label{sec:real}}
\subsection{Financial data \label{sec:financial}}
\begin{figure}
\includegraphics[width=3.38in]{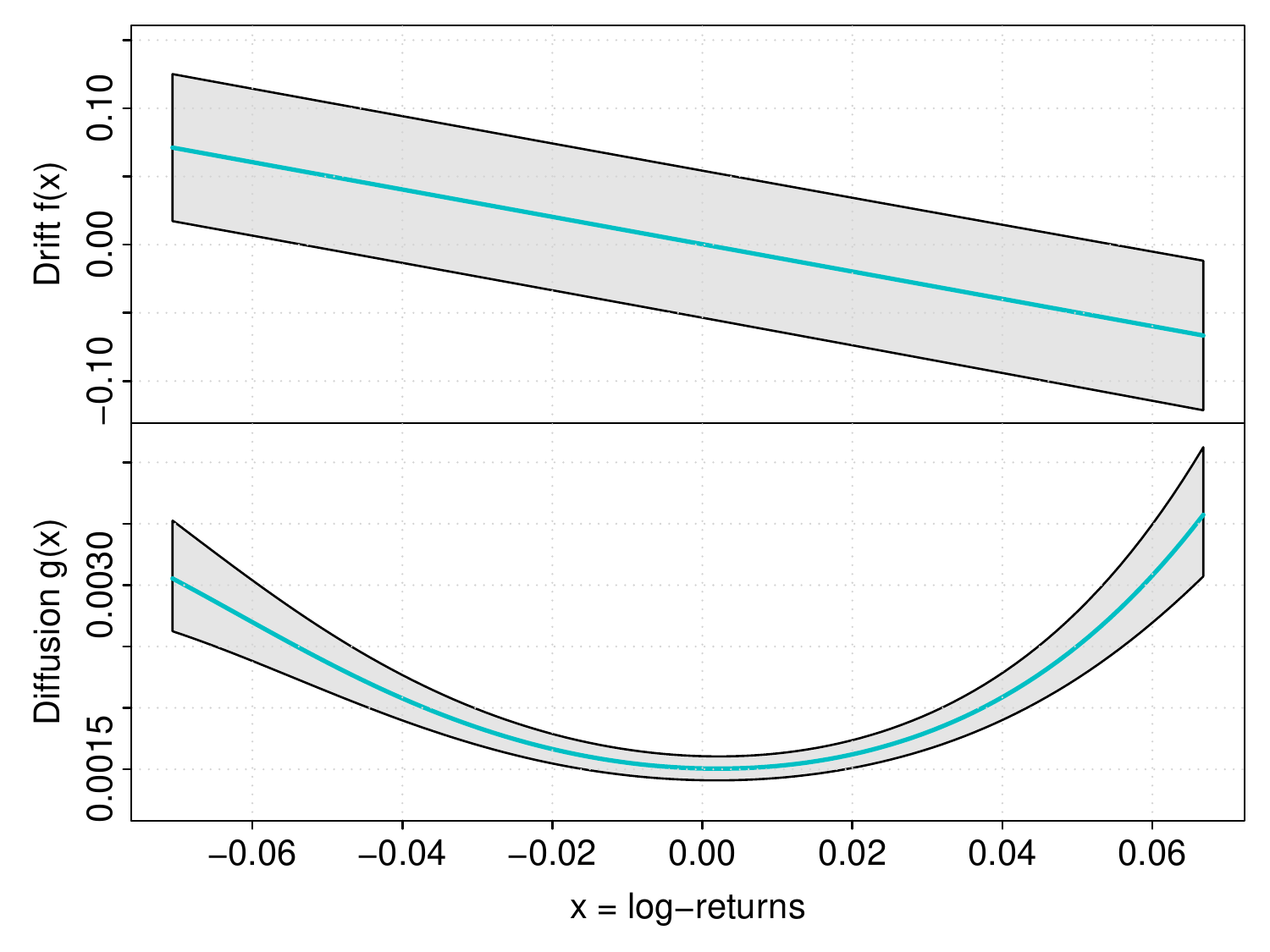} 
\caption{Drift and diffusion estimates obtained with the SGP method on the oil 
price log-returns.\label{fig:oil_estimates}}
\end{figure}

In this Section, we apply our method to a real time series from econophysics 
with the aim of illustrating the applicability of SDEs to non-stationary
problems and the role that non-constant diffusions play in complex dynamics.
We study the daily fluctuations in the oil price in the period 
1982/01/02-2017/05/30, which results in a time series $\bm{p}$ of length 
$N \approx 10^4$ \cite{eiaData}. Following \cite{ghasemi2007markov}, we 
constructed the daily 
logarithmic increments of the oil price $x_n = \log p_{n + 1} / p_n$ to obtain a
stationary time series. The SGP method was then applied using $m=10$ inducing 
points (randomly started using the percentile initialization) and two squared
exponential kernels. The numerical stability of the kernels was improved using
Eq.~\eqref{eq:modified_kernel}. The amplitudes of the kernels 
were selected to match a standard deviation of 5 for both the drift and 
diffusion functions. The algorithm was run several times with 
random initial values for the length-scales. The final estimates
selected using the lower bound are shown in Fig.~\ref{fig:oil_estimates}. These
estimates are in good agreement with those reported in \cite{ghasemi2007markov}
(although this work focused in a smaller period). Similar estimates are
also obtained using the \textit{KBR} and \textit{POLY} methods. 

The drift and diffusion functions shown in Fig.~\ref{fig:oil_estimates} can 
be approximated by $\hat{f}(x) \approx -x$ and 
$\hat{g}(x) \approx D + \gamma x^2$, which yields the SDE of a
quadratic-noise Ornstein-Uhlenbeck process 
\cite[Chapter~3]{gillespie1991markov}. This process is an illustrative example
of the effects that multiplicative noise may have in the dynamics of a system. 
The stationary distribution of a quadratic-noise Ornstein-Uhlenbeck process
is a non-standardized Student's distribution, which is a heavy-tailed
distribution that permits the occurrence of large values in the log-returns
series. Furthermore, this stationary distribution is 
more closely confined to the origin in comparison with the standard 
Ornstein-Uhlenbeck noise, which implies that the stable state is narrower in 
the quadratic case. This is an example of noise-enhanced stability 
\cite[Chapter~3]{gillespie1991markov} and illustrates the importance that the 
non-parametric estimation of the diffusion may have in the study of complex 
dynamics.
  
\subsection{Paleoclimatology data \label{sec:paleoclimatology}}
In this Section, we apply our estimation algorithm to a real data problem 
related to paleoclimatology. Climate records from the Greenland ice cores have
played a central role in the study of the Earth's past climate in the Northern
hemisphere. Among other  interesting phenomena, these records show 
abrupt rapid climate  fluctuations that occurred during the 
last glacial period, which ranges from approximately 110 Ky (1 Ky = 1000 years)
to 12 Ky before present. These abrupt climate changes are 
usually referred to as  Dansgaard-Oeschger (DO) events. Although there seems to
be a general agreement that DO events are transitions between two 
quasi-stationary states (the glacial or stadial and the interstadial 
states), it is still actively debated the nature of the phenomena triggering the
transitions. It has been argued that the DO events occur quasi-periodically with
a recurrence time of approximately 1.47 Ky \cite{schulz20021470}. However,
recent studies support that the DO events  are probably noise induced 
\cite{ditlevsen2007climate,ditlevsen2009stochastic,krumscheid2015data}. 

We apply our method to the $\delta^{18}\text{O}$ record during 
the last glacial period obtained from the North Greenland Ice Core Project
(NGRIP) \cite{andersen2004high}. The $\delta^{18}\text{O}$ is a measure of the 
ratio of the stable isotopes oxygen-18 and oxygen-16 which is commonly used
to estimate the temperature at the time that each small section of the ice core
was formed. It is measured in ``permil" (\permil, parts per thousand) and its 
formula is:
\begin{equation*}
\delta^{18}\text{O} = \Bigg( \frac{\left[
\frac{^{18}\text{O}}{^{16}\text{O}}\right]_{\text{sample}}}
{\left[\frac{^{18}\text{O}}{^{16}\text{O}}\right]_{\text{reference}}} - 1\Bigg) 
\cdot 1000 \text{ \permil},
\end{equation*}
where \textit{reference} defines a well-known isotopic composition.

\begin{figure}
\includegraphics[width=3.38in]{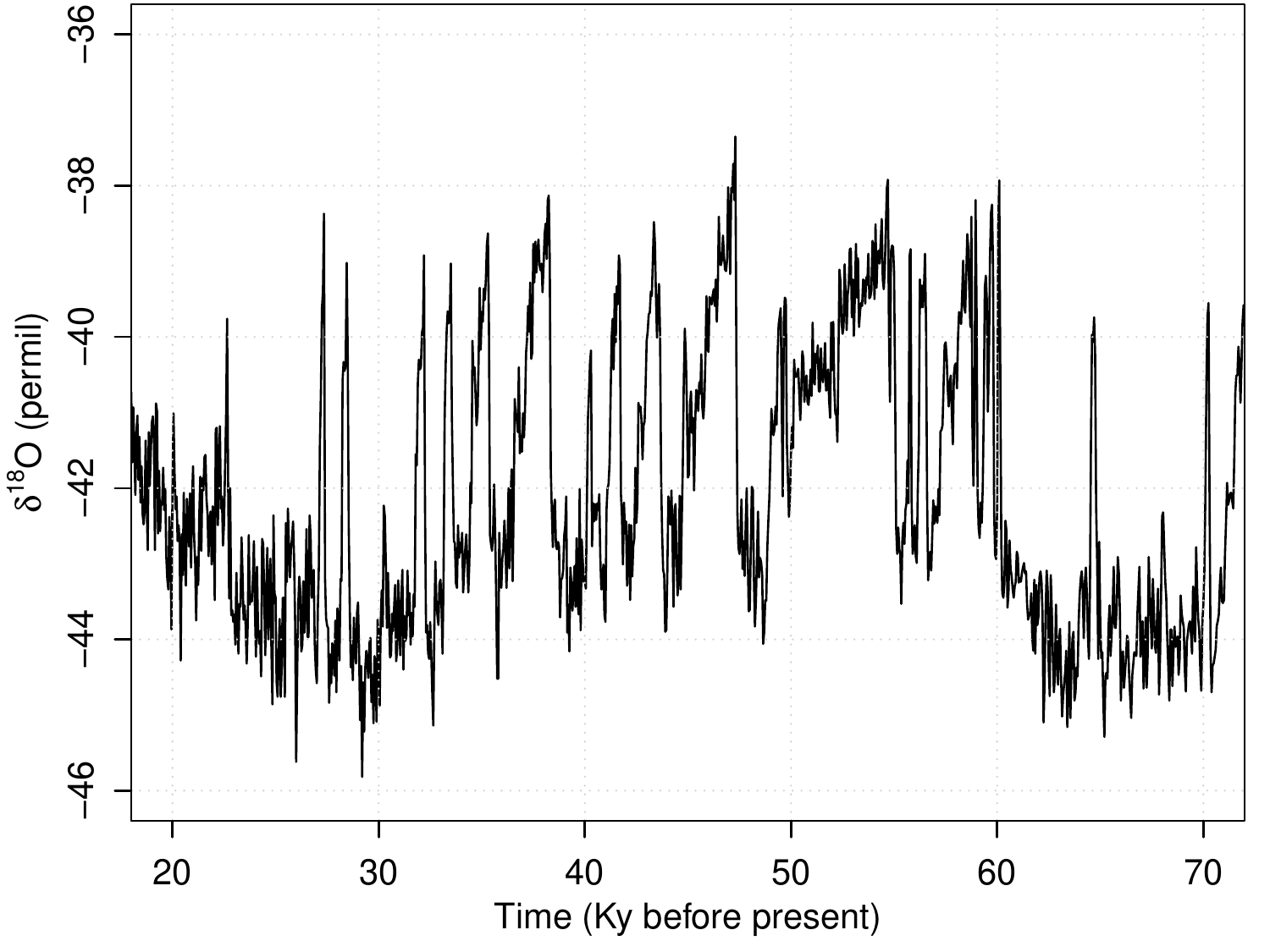} 
\caption{DO transitions during 
the last glacial period.}\label{fig:do_transitions}
\end{figure}

Fig.~\ref{fig:do_transitions} shows the oxygen isotopic composition from the
NGRIP ice core. We consider the period ranging from 70 Ky to 20 Ky before
present as in \cite{krumscheid2015data}, since it is dominated by the DO events, 
as can be clearly observed.

We applied our method using different kernels to illustrate that different 
covariance functions can be used and combined to create different models, and 
that Eq.~\eqref{eq:L_comparison} can be used to select the best among them. 
Within our method, testing different kernels is important because we usually do 
not have enough information about the drift and diffusion terms to decide among 
them. Furthermore, the performance of GPs depends almost exclusively on the
suitability of the chosen kernel to capture the features of the modelled 
function. Consider the following illustrative example: a function 
with fast quasi-periodic oscillations superimposed on a linear trend. A squared 
exponential kernel with a large length-scale can capture the behaviour of the 
linear slope and make reasonable predictions of the trend for unobserved values,
but it won't be able to model the quick wiggles. On the other hand, a squared 
exponential kernel with a small length-scale will be able to accurately fit all
the data but, since the distance from the training points rapidly increases, it
won't be able to make good predictions for unobserved values, not even for the 
trend. Furthermore, the uncertainty of the unobserved values will also scale 
fast. A better covariance choice could make use of a sum of exponential kernels
with different length-scales, which would permit to accurately fit the data and
make good predictions for the trend. More complex kernel choices are also 
possible. For a complete example on the impact of the kernel in the modelling
capabilities of a GP, see \cite[Chapter~5]{rasmussen2006gaussian}. For our 
illustrative example on the paleoclimate data, we used the kernel specified in 
Eq.~\eqref{eq:general_kernel}, a sum of two exponential kernels with different
length-scales and a rational quadratic kernel. All 
these kernels were modified adding a small value to their main diagonals as in 
Eq.~\eqref{eq:modified_kernel}. 

\begin{figure*}
\subfigure{\label{fig:do_estimates_kbr}}
\subfigure{\label{fig:do_estimates_sgp}}
\subfigure{\label{fig:do_estimates_poly}}
\includegraphics[width=6.5in]{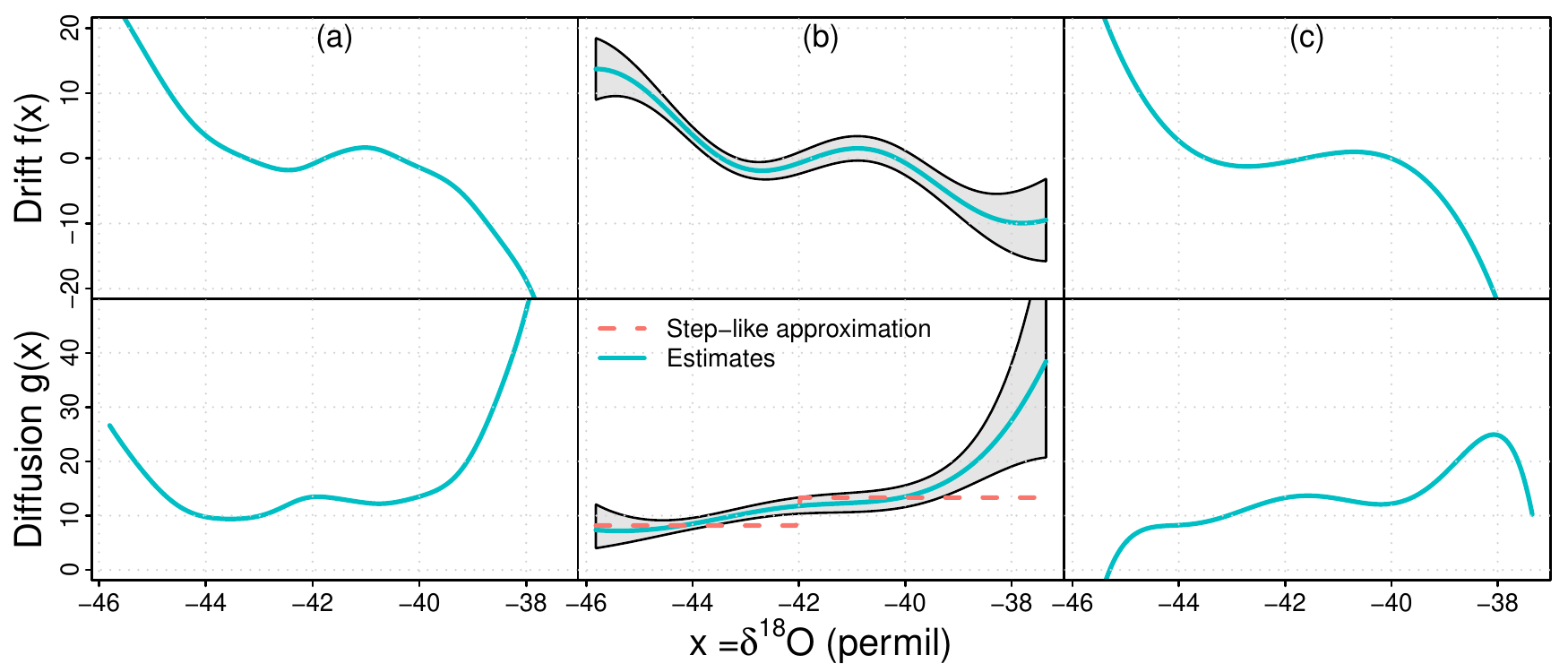} 
\caption{Best drift and diffusion estimates using the (a) KBR, 
(b) SGP and (c) POLY methods with the paleoclimate data.}
\label{fig:do_estimates}
\end{figure*}

For each possible kernel, the method was started with random values for the
hyperparameters. The number of the pseudo-inputs was set to $m=15$, based on the
good results that it achieved at Section \ref{sec:synthetic}. The amplitudes of 
the kernels were  selected so that they were compatible with a standard 
deviation of 30 for both the drift and diffusion functions. The estimates
selected based on the value of the modified lower bound 
(Eq.~\eqref{eq:L_comparison}) are illustrated
in Fig.~\ref{fig:do_estimates_sgp}. The drift term was obtained
using a rational  squared kernel whereas the diffusion term was estimated using
the kernel from  Eq.~\eqref{eq:general_kernel}.  Note that, as expected, the 
drift function  presents two stable points: one  corresponding to the stadial
state and the  other corresponding to the interstadial state. Integrating the
drift function  yields the potential function, which indicates
that the stadial state corresponds to a stable state of the system since it has
the lowest energy. On the other hand, the interstadial state corresponds to a
metastable state. 

The SGP estimate supports the use of a state-dependent diffusion 
rather than the widely-used constant term. The use of a state-dependent 
diffusion for the DO events was first proposed in \cite{krumscheid2015data},
which suggested:
\begin{equation}
f(x, \bm{\theta}) = \sum_{i = 0}^{3} \theta_j x^j; \qquad  g(x, \bm{\theta}) = 
            \begin{cases}
                \theta_4 \text{ if } x < \theta_6\\
                \theta_5 \text{ if } x \geq \theta_6
            \end{cases}.
\label{eq:krumModel}
\end{equation}

Krumscheid et al. suggested the model from Eq.~\eqref{eq:krumModel} while 
testing their framework for parametric inference and model selection for SDEs 
\cite{krumscheid2015data}. The authors
discussed the model from Eq.~\eqref{eq:krumModel} since it is able to 
accurately predict the histogram of the DO events, although the final
parametrization that results from their model selection criteria
proposes a constant diffusion term. However, our
non-parametric methodology suggests that the state dependent diffusion
is indeed preferable. Despite it is possible to approximate the SGP diffusion's 
estimate using a step function (as can be appreciated in
Fig.~\ref{fig:do_estimates}), there exists a linear increasing region for 
$x > -40$ that does not match Krumscheid's
model. To compare the diffusion model in Ref.~\cite{krumscheid2015data}
with the SGP's diffusion model, a Lasso estimate \cite{de2012adaptive} was
applied to the  diffusion term  while keeping  the drift term fixed. Lasso 
penalties are very useful in regression analysis since they are able to set 
coefficients to zero, eliminating unnecessary variables.  Hence, by using the 
diffusion term:
\begin{equation}
g(x, \bm{\theta}) = 
            \begin{cases}
                \theta_1 \text{ if } x < -42\\
                \theta_2 \text{ if } -42 \leq x < -40\\
                \theta_2 + \theta_3 (x + 40) \text{ if } x \geq -40
            \end{cases},
\end{equation}
we can compare the model proposed in \cite{krumscheid2015data} (which
corresponds with setting  $\theta_3 = 0$) and our estimate. The Lasso 
estimate provides evidence in favour of the model obtained through our method,
since the $\theta_3$ is not eliminated. Note, however, that this
evidence it is not conclusive since the time series used for the estimates is 
quite short  ($N \approx 10^3$) and there is a lack of points for 
$x > -39$.

We have also used the SGP estimates to compute the distribution 
of the time between DO events. 1000 new time series were generated using
the Euler-Maruyama scheme. The initial points were sampled with replacement from
the real paleoclimate series. To robustly  identify the  DO states, we fitted a
Hidden Markov Model (HMM) with three states and Gaussian response to the real 
data. The aim of the three states is to clearly identify the stadial state, the
interstadial state and a ``transition state". We identified the states of each
of the simulated time series using this HMM by means of the Viterbi algorithm 
\cite{viterbi1967error}. The resulting mean time between DO events was
1.50 Ky, in good agreement with 
the generally accepted value of 1.47 Ky \cite{schulz20021470}. However, it must 
be noted that this value was obtained on basis of a quasi-periodical model, 
whereas our value is based on a stochastic model.

Using the KBR (Fig.~\ref{fig:do_estimates_kbr}) and
POLY (Fig.~\ref{fig:do_estimates_poly}) methods
result in similar drift estimates, compared with the SGP one. Furthermore, both
diffusion estimates also support the use of a the state-dependent model. 
However, the SGP estimate provides confidence intervals based on a Bayesian 
setting while the others methods do not. Also, the  KBR method presents an 
unlikely increment in the diffusion for $x < -44$. The POLY method approximates 
the shape of the state-dependent diffusion using a high order polynomial, which 
cannot  properly capture the plateau for $x < -44$. Additionally, the polynomial
fit results in negative values for $x < -45$ which have no sense for a diffusion
term.

\section{Discussion and conclusions \label{sec:conclusion}}
In this paper, we have presented a non-parametric estimation method for
SDEs from densely-observed time series based on GPs. The only assumptions
made on the data are that they fulfil the Markovian condition and that the 
sampling period is small enough so that the Euler-Maruyama discretization holds.
From the point of view of the adoption of GPs to the estimation of SDEs, the 
main contributions of this paper are: (1) providing estimates for any type of
diffusion function and (2) 
proposing a sparse approximation to the true GP posterior that permits to
efficiently handle the typical experimental time series size of 
$N \approx 10^3-10^5$. To cope with the computational complexity of calculating
the posterior distribution of the GPs (which scales as  $\mathcal{O}(N ^ 3)$), 
we approximate the GPs using the evidence provided by the data in only a small
set of function points, the inducing variables. The inducing  variables are 
learnt by minimizing the Kullback-Leibler divergence between the true posterior 
GP distribution and the approximated one. The minimization problem is approached
using the standard techniques from the variational inference framework, which
usually yields a coordinate-ascent optimization to approximate the posterior. 
However, our approach makes use of a non conjugate model due to the inclusion
of the diffusion function, which prevents the direct use of
variational inference methods. To tackle the problem, a Laplace approximation 
was used to compute the distribution modelling the diffusion.  It must be noted
that, although we have developed our estimation approach bearing in mind the
computational challenges that a large $N$ imposes, our proposal can also handle 
small time series without any further adjustment. Also, although the SGP 
approximation permits handling large experimental time series, $N$ cannot 
increase without limit. Variational inference algorithms require a full pass
through the whole dataset at each iteration and hence, they become inefficient 
for massive datasets, even when using sparse techniques like the proposed one. 
For example, in a computer with an Intel Xeon E5-2650L at 2.05 GHz and 
using $m=10$, the computation time increases from 8 minutes per iteration 
with $N=10 ^ 5$ to 1.4 hours per iteration with $N=10^6$. Scaling up variational
inference can be done
using stochastic gradient optimization, which yields stochastic variational 
inference \cite{hoffman2013stochastic}. Since large datasets are increasingly 
common, the use of this kind of techniques should be considered in future work.

The performance of the SGP estimates was evaluated using simulated data from
different SDE models and compared with the kernel based method 
\cite{lamouroux2009kernel} and the polynomial based method 
\cite{rajabzadeh2016robust}. The results show that the SGP approach is
able to provide very accurate estimates, specially for the drift term. The main
advantage of the SGP method with respect to 
\cite{lamouroux2009kernel} and \cite{rajabzadeh2016robust} is that it permits a
Bayesian treatment of the estimation problem; this enables obtaining 
probabilistic predictions and computing robust confidence intervals. 
Furthermore, the prior information 
about the drift/diffusion is expressed in a function-space view, i.e.,
the SGP method permits specifying the prior directly over functions instead of
working with weights of some basis expansion. In our view, this is a more
natural way of working with functions. Another major advantage of the proposed
method is its versatility. Although we have focused on very flexible kernels, 
any type of kernel (or even combinations of them) can be used, which may
completely change the properties of the posterior estimates. For example, using 
polynomial kernels would yield  similar estimates to those of
\cite{rajabzadeh2016robust}, but with the aforementioned advantages of the 
Bayesian framework and without the possibility of obtaining
negative values for the diffusion (see Section~\ref{sec:real}).

We applied the SGP method to a real problem in econophysics with the aim 
of illustrating the importance of non-constant diffusions in the behaviour
of a system and, hence, the importance of its non-parametric estimation. This
example also emphasizes the applicability of the SDE framework to non-stationary
time series.

The proposed method was also applied to a real paleoclimate time series: the 
NGRIP core data showing the DO events occurring during the last glacial period.
The SGP method accurately captures the relevant physical states of the time 
series (the  stadial and interstadial states) and yields a mean transition time 
between DO events that it is close to the accepted value in the literature
under the assumption  of a deterministic periodic model. This demonstrates 
its ability to capture the behaviour of real data with complex dynamics. 
Furthermore, the SGP estimates  provide evidence supporting a novel state 
dependent diffusion model for the DO events. This diffusion model is similar
to the step-like function  proposed in \cite{krumscheid2015data} for a wide
range of the diffusion's support, but it also adds a linear term for the 
region corresponding to the most extreme values of the DO events. These 
results should be viewed with caution, since the estimates were made using
small amounts of data. Further research to assess the physical meaning of the
model should be made. 

In future work we would like to design some criteria to automatically 
estimate an  appropriate number of pseudo-inputs $m$ taking into account both
the modified lower  bound (Eq.~\eqref{eq:L_comparison}) and the additional 
computational time  required when $m$ increases. The substitution of the 
exponential transformation used to ensure the positiveness of $g$ in favour of 
another less-explosive  transformation should also be considered in future 
improvements. The numerical stability of the method would certainly improve with
the use of smoother transformations, but the formal expressions required for the 
variational inference problem would be more complicated. An alternative to avoid
the complicated mathematical expressions would be to use black-box variational 
inference frameworks \cite{NIPS2015_5678}. Since black-box methods are based in 
stochastic variational inference, its application would also permit to scale the
exposed methodology to datasets much bigger that those studied in this article.
Hence, The application of black-box methods to the reconstruction of SDEs 
looks promising and should be explored in future work. 

We believe that the presented method could help to further comprehend the
dynamics  underlying a wide variety of complex systems. To that end, we provide 
an open-source implementation of our method which is freely available at github 
(\githubUrl).

\begin{acknowledgments}
This work has received financial support from the Conseller\'ia de Cultura, 
Educaci\'on e Ordenaci\'on Universitaria (accreditation 2016-2019, ED431G/08),
the European Regional Development Fund (ERDF), by the   Spanish   MINECO 
under the project  TIN2014-55183-R and by San Pablo CEU University under the 
grant PCON10/2016. Constantino A. Garc\'{i}a  acknowledges the
support of the FPU Grant program from the Spanish Ministry of Education (MEC) 
(Ref. FPU14/02489). 
\end{acknowledgments}

\appendix
\section{Variational distribution for the drift \label{ap:a}}
We start expanding the expression inside the expectation operator from Equation
\eqref{eq:vi_sol_f}:
\begin{equation*}
\begin{split}
\log \phi_{f_m}(\bm{f}_m) =& \mathbb{E}_{\phi_{\uminus f_m}}\left[ 
\log \left(p(\bm{x}|\bm{f}, \bm{s})p(\bm{s}_m)p(\bm{f}_m)\right)\right] \\
=&\mathbb{E}_{\phi_{\uminus f_m}}\left[ \log p(\bm{x}|\bm{f}, \bm{s})\right] 
+\mathbb{E}_{\phi_{\uminus f_m}}\left[\log p(\bm{s}_m)\right]\\
&+ \mathbb{E}_{\phi_{\uminus f_m}}\left[\log p(\bm{f}_m)\right].
\end{split}
\end{equation*}
Given that the expectation operator does not affect $\log p(\bm{f}_m)$ 
and that when applied to $\log p(\bm{s}_m)$ results in an 
expression that does not depend on $\bm{f}_m$, we may write:
\begin{equation}
\begin{split}
\log \phi_{f_m}(\bm{f}_m) =&\mathbb{E}_{\phi_{\uminus f_m}}\left[
\log p(\bm{x}|\bm{f}, \bm{s})\right]\\ 
&- \frac{1}{2}\bm{f}_m^T  \bm{K}_{mm}^{-1}\bm{f}_m + \text{constant},
\label{eq:f_step1}
\end{split}
\end{equation}
where we have denoted all terms that do not depend on $\bm{s}_m$ as
\textit{constant}. It is convenient to work with the term \textit{constant},
given that we can infer its value after identifying the distribution of
$\phi_{f_m}$. In that case, \textit{constant} corresponds to the normalizing 
constant required by  the distribution $\phi_{f_m}$. 

Expanding the expression inside the expectation operator
from Eq.~\eqref{eq:f_step1} and joining new constants yields:
\begin{equation}
\begin{alignedat}{3}
\log \phi_{f_m}(\bm{f}_m) =&-\frac{1}{2\Delta t}\sum_{i=1}^N
&&\mathbb{E}_{\phi_{s_m}(\bm{s}_m) p(\bm{s}\mid \bm{s}_m)}\left[\exp(-s_i)
\right] \\ 
& &&\times \mathbb{E}_{p(\bm{f}\mid \bm{f}_m)}\left[(\Delta x_i
- \Delta tf_i)^2\right]\\ 
&- \frac{1}{2}\bm{f}_m^T \bm{K}_{mm}^{-1} &&\bm{f}_m + \text{constant}.
\end{alignedat}
\label{eq:f_step2}
\end{equation}
Using Fubini's rule of integration we may write the first 
expectation from Eq.~\ref{eq:f_step2} as:
\begin{equation}
\label{eq:f_step3}
\mathbb{E}_{\phi_{s_m}(\bm{s}_m) p(\bm{s}\mid \bm{s}_m)}\left[\exp(-s_i) 
\right] =
\mathbb{E}_{\phi_{s_m}}\big[\mathbb{E}_{p(\bm{s}\mid \bm{s}_m)}
\left[\exp(-s_i)\right]\big].
\end{equation}

It is possible to demonstrate that if $X \sim \mathcal{N}(\mu, \sigma ^2)$, 
$\mathbb{E}\left[\exp(-X)\right] = \exp(-\mu + \sigma ^2 / 2)$. Hence, 
Eq.~\eqref{eq:f_step3} becomes:
\begin{equation}
\begin{split}
\label{eq:f_step4}
\mathbb{E}_{\phi_{s_m}}&\big[\mathbb{E}_{p(\bm{s}\mid \bm{s}_m)}
\left[\exp(-s_i)\right]\big]=\\
&\mathbb{E}_{\phi_{s_m}} \left[-[\bm{v}^N + \bm{B}(\bm{s}_m - 
\bm{v}^m)]_i\right] + \frac{Q_{ii}}{2} = 
\zeta_i,
\end{split}
\end{equation}
where we have used the definition of $\zeta_i$ from Eq.~\eqref{eq:zeta}
and the definitions of $\textbf{B}$ and $\textbf{Q}$ from 
Eq.~\eqref{eq:aux_matrices}. Introducing back Eqs.~\eqref{eq:f_step3} and 
\eqref{eq:f_step4} into Eq.~\eqref{eq:f_step2} we finally arrive to:

\begin{widetext}
\begin{align}
\log \phi_{f_m}(\bm{f}_m) &=-\frac{1}{2\Delta t}\sum_{i=1}^N \nonumber
\zeta_i\cdot\mathbb{E}_{p(\bm{f}\mid \bm{f}_m)}\left[(\Delta x_i) ^ 2
-2\Delta  t\Delta x_i f_i - (\Delta t)^2f_i^2\right]
- \frac{1}{2}\bm{f}_m^T \bm{K}_{mm}^{-1} \bm{f}_m + \text{constant}\\ 
&=-\frac{1}{2\Delta t }\sum_{i=1}^N
\zeta_i\cdot\left[(\Delta x_i) ^ 2
- 2\Delta t\Delta x_i [\bm{A}\bm{f}_m]_i
-(\Delta t)^2([\bm{A}\bm{f}_m]_i^2 + P_{ii})\right]
- \frac{1}{2}\bm{f}_m^T \bm{K}_{mm}^{-1}\bm{f}_m + \text{constant}
\label{eq:f_step5}
\end{align}
\end{widetext}

Reordering Eq.~\eqref{eq:f_step5} an expressing it in vectorial form, we 
recover Eq.~\eqref{eq:f_sol}.

\section{Variational distribution for the diffusion \label{ap:b}}
Starting from Eq.~\eqref{eq:vi_sol_s} and proceeding similarly to 
Appendix \ref{ap:a} it is possible to arrive to:

\begin{equation}
\begin{alignedat}{3}
\log &\phi_{s_m}(\bm{s}_m) =& & \\
&-\frac{1}{2}(\bm{s}_m - && \bm{v}^m)^T \bm{J}_{mm}^{-1}(\bm{s}_m - \bm{v}^m) \\
&-\frac{1}{2\Delta t}\sum_{i=1}^N &&
\mathbb{E}_{p(\bm{s}\mid \bm{s}_m)}\left[\exp(-s_i) \right]\\
& &&\times \mathbb{E}_{\phi_{f_m}(\bm{f}_m)p(\bm{f}\mid \bm{f}_m)}
\left[(\Delta x_i -
\Delta tf_i)^2\right]\\
&-\frac{1}{2} \sum_{i=1}^N && \mathbb{E}_{p(\bm{s}\mid \bm{s}_m)}
\left[ s_i \right] +
\text{constant}.
\end{alignedat}
\label{eq:s_step_half}
\end{equation}

The expectation of $\exp(-s_i)$ can be computed as 
in Appendix \ref{ap:a} (see Eq.~\eqref{eq:f_step3}), which results in
\begin{equation}
\begin{alignedat}{3}
\log & \phi_{s_m}(\bm{s}_m) = &&\\
&-\frac{1}{2}(\bm{s}_m - &&\bm{v}^m)^T \bm{J}_{mm}^{-1}(\bm{s}_m - \bm{v}^m)\\
&-\frac{1}{2\Delta t}\sum_{i=1}^N &&
\exp\bigg(-[\bm{v}^N + \bm{B}(\bm{s}_m - \bm{v}^m)]_i +\frac{Q_{ii}}{2}\bigg)\\
& && \times \mathbb{E}_{\phi_{f_m}(\bm{f}_m)p(\bm{f}\mid \bm{f}_m)}
\left[(\Delta x_i - \Delta tf_i)^2\right]\\
&-\frac{1}{2}\sum_{i=1}^N &&[\bm{B}(\bm{s}_m - \bm{v}^m)]_i+ \text{constant}.
\end{alignedat}
\label{eq:s_step_1}
\end{equation}

Using again Fubini's law, we may write:
\begin{align*}
\mathbb{E}_{\phi_{f_m}(\bm{f}_m)p(\bm{f}\mid \bm{f}_m)}
&\left[(\Delta x_i - \Delta tf_i)^2\right] \\
= \mathbb{E}_{\phi_{f_m}}&\left[\mathbb{E}_{p(\bm{f}\mid
\bm{f}_m)}\left[(\Delta x_i - \Delta tf_i)^2\right]\right]&\\
=\mathbb{E}_{\phi_{f_m}} &
\left[\Delta x_i^2 \right. - 2 \Delta t\Delta x_i [\bm{A}\bm{f}_m]_i &\\
 &\left. + (\Delta t)^2\left([\bm{A}\bm{f}_m]_i^2 + P_{ii}\right) \right]
= \psi_i, &
\end{align*}
where we have used the definition of $\psi_i$ from Eq.~\eqref{eq:psi}. 
Introducing $\psi_i$ into Eq.~\eqref{eq:s_step_1}, we finally arrive to 
Eq.~\eqref{eq:s_sol}.

\vspace{1mm}
\bibliography{bibliography}
\end{document}